\documentclass{article}

\PassOptionsToPackage{numbers,sort,compress}{natbib}
\usepackage[final]{neurips_2025}

\usepackage{amsmath,amsfonts,bm}

\def\eqref#1{equation~\ref{#1}}

\def\1{\bm{1}}

\DeclareMathAlphabet{\mathsfit}{\encodingdefault}{\sfdefault}{m}{sl}
\SetMathAlphabet{\mathsfit}{bold}{\encodingdefault}{\sfdefault}{bx}{n}

\usepackage[utf8]{inputenc}

\usepackage{hyperref}
\usepackage{url}
\usepackage{algorithm}
\usepackage{algpseudocodex}
\usepackage{amsmath}
\usepackage{amssymb}
\usepackage{amsfonts}
\usepackage{amsthm}
\usepackage{bbm}
\usepackage{dsfont}
\usepackage{multicol}
\usepackage{multirow}
\usepackage{threeparttable}
\usepackage{tempora}
\usepackage{wrapfig}
\usepackage{subcaption}

\usepackage{booktabs}       %
\usepackage{nicefrac}       %
\usepackage{microtype}      %
\usepackage{xcolor}         %

\usepackage[T1]{fontenc}
\usepackage[russian,english]{babel}

\newtheorem{problem}{Problem}

\title{Approximately Aligned Decoding}

\author{Daniel Melcer\thanks{Work performed while at Amazon} \\
Khoury College of Computer Sciences\\
Northeastern University\\
Boston, MA, USA \\
\texttt{melcer.d@northeastern.edu} \\
\And
Sujan Gonugondla$^*$ \\
Meta Superintelligence Labs \\
New York, NY, USA \\
\texttt{sujan@meta.com} \\
\And
Pramuditha Perera \\
AWS NGDE \\
New York, NY, USA \\
\texttt{pramudi@amazon.com} \\
\And
Haifeng Qian$^*$ \\
Nvidia \\
Santa Clara, CA, USA \\
\texttt{haifengq@nvidia.com} \\
\And
Wen-Hao Chiang \\
AWS NGDE \\
New York, NY, USA \\
\texttt{cwenhao@amazon.com} \\
\And
Yanjun Wang \\
AWS NGDE \\
New York, NY, USA \\
\texttt{yanjunw@amazon.com} \\
\And
Nihal Jain \\
AWS NGDE \\
New York, NY, USA \\
\texttt{nihjain@amazon.com} \\
\And
Pranav Garg \\
AWS NGDE \\
New York, NY, USA \\
\texttt{prangarg@amazon.com} \\
\And
Xiaofei Ma \\
AWS NGDE \\
New York, NY, USA \\
\texttt{xiaofeim@amazon.com} \\
\And
Anoop Deoras \\
AWS NGDE \\
New York, NY, USA \\
\texttt{adeoras@amazon.com} \\
}

\algnewcommand{\algorithmicwithprob}{\textbf{with probability}}

\makeatletter
\algdef{SE}[WITHPROB]{WithProb}{EndWithProb}[1]{%
	\algpx@startCodeCommand\algpx@startIndent\algorithmicwithprob\ #1\ \algorithmicdo%
}{%
	\algpx@startEndBlockCommand\algpx@endIndent\algorithmicend\ \algorithmicwithprob%
}
\algdef{SE}[WITHPROB]{StandaloneElse}{EndStandaloneElse}[0]{%
	\algpx@startCodeCommand\algpx@startIndent\algorithmicelse%
}{%
	\algpx@startEndBlockCommand\algpx@endIndent%
}

\pretocmd{\EndWithProb}{\algpx@endIndent}{}{}%
\pretocmd{\EndStandaloneElse}{\algpx@endIndent}{}{}%

\pretocmd{\WithProb}{\algpx@endCodeCommand}{}{}
\pretocmd{\StandaloneElse}{\algpx@endCodeCommand}{}{}

\pretocmd{\EndWithProb}{\algpx@endCodeCommand[1]}{}{}%
\pretocmd{\EndStandaloneElse}{\algpx@endCodeCommand[1]}{}{}%

\algtext*{EndWithProb}
\algtext*{EndStandaloneElse}

\def\blfootnote{\xdef\@thefnmark{}\@footnotetext}

\makeatother

\begin{document}

\newcommand{\badSet}{\mathcal{B}}
\newcommand{\llm}{P}
\newcommand{\llmExcBadSet}{\hat{\llm}^\badSet}
\newcommand{\badSeen}{B}
\newcommand{\llmExcBadSeen}{\hat{\llm}^\badSeen}
\newcommand{\llmExc}[1]{\hat{\llm}^{#1}}

\newcommand{\llmExcBadSeenAndX}{\hat{\llm}^{\badSeen \cup \{x\}}}

\newcommand{\llmExcBadSetSample}{\hat{\llm}^\badSet_\alpha}

\newcommand{\ssm}{\llm_{\textit{Spec}}}

\newcommand{\constrainedGenerationSampler}{\mathcal{C}^\badSet_\llm}

\newcommand{\aadSampler}{\mathcal{A}^\badSet_\llm}

\newcommand{\vocab}{\mathcal{V}}

\newcommand{\indicator}[1]{\mathbbm{1}_{#1}}

\newcommand{\genBox}[3]{%
\fbox{\parbox{0.98\linewidth}{#1\hfill #2

\vspace{0.2em}
\centering
\noindent\fbox{
\parbox{0.96\linewidth}{#3}}}}}

\newcommand{\rus}[1]{\textcolor{red}{\underline{\foreignlanguage{russian}{#1}}}}

\maketitle

\begin{abstract}
It is common to reject undesired outputs of Large Language Models (LLMs); however, current methods to do so require an excessive amount of computation to re-sample after a rejection, or distort the distribution of outputs by constraining the output to highly improbable tokens.
We present a method, Approximately Aligned Decoding (AprAD), to balance the distortion of the output distribution with computational efficiency, inspired by algorithms from the speculative decoding literature.
AprAD allows for the generation of long sequences of text with difficult-to-satisfy constraints, while amplifying low probability outputs much less compared to existing methods.
We show through a series of experiments that the task-specific performance of AprAD is comparable to methods that do not distort the output distribution, while being much more computationally efficient.
\end{abstract}

\section{Introduction}

Large Language Models (LLMs) are able to perform many complex text manipulation tasks, and embody an incredible amount of world knowledge, but their output may contain undesirable elements such as syntactically-incorrect code, hallucinated PII, profanity, or failed tool calls.
These issues, which we collectively refer to as errors or constraint violations for the remainder of the paper, may be detected with incremental parsers, regular expression matching, or even simple substring searches.\blfootnote{Code available at \url{https://github.com/amazon-science/Approximately-Aligned-Decoding}.}
\setcounter{footnote}{0}

Each individual task that a LLM is used for may have a unique set of constraints. 
However, re-training a LLM to accommodate the constraints of every task is expensive, and may still not fully protect against violations.
Therefore, the community has developed several methods to mitigate constraint violations without the need to retrain the language model. 
However, many existing methods deviate severely from the original output distribution, or have intractable performance for certain constraints.

Our contributions are as follows.
First, we analyze several existing methods for avoiding constraint violations in text generated from autoregressive language models, and compare the strengths and weaknesses of each method.
Second, we present a method, Approximately Aligned Decoding (AprAD), that allows for a useful midpoint in the tradeoff between computational efficiency and maintenance of the output distribution, without the need for any additional training or fine-tuning step. 
At its core, AprAD uses a procedure from the speculative sampling literature to determine backtracking behavior after encountering an error generation.
Finally, we run a series of experiments, demonstrating that our method obtains excellent task-specific performance on both synthetic and real-world domains, without introducing an unreasonable level of inference overhead.

\subsection{Related Work}

Language models based on a Transformer architecture \citep{vaswani2023attentionneed} have steadily become more popular, with high parameter counts, in consumer chatbot products such as OpenAI ChatGPT \citep{OpenAI_2024} and Anthropic Claude \citep{Claude_2024}, or code generation tools such as GitHub Copilot \citep{githubinc.GitHubCopilotYour2023} and Amazon Q Developer \citep{amazonwebservicesinc.AICodeGenerator2023}.

\textbf{Sampling-based methods.}
Several LLM tools have introduced output control features such as schema-restricted generation \citep{OpenAI_2024b}.
For those willing to run local inference on a language model, however, there are a vast array of methods for constraining the output of a model to follow a template \citep{microsoftTypeChat2023,sengottuveluJsonformerBulletproofWay2023,automorphicTrex2023, microsoftGuidance2023,sriLQML2023,athiwaratkun2024tokenalignmentcharactermatching}, produce syntactically valid code \citep{jonesLlamaAddGrammarbased2023, slattonAddedContextFree2023,willard2023efficient,takerngsaksiriSyntaxAwareOntheFlyCode2023,melcer2024constraineddecodingcodelanguage}, or conform to various poetry constraints \citep{roush2023languagemodelspoetstoo}.
These works often use \textit{constrained generation} \citep{beurerkellner2024guidingllmsrightway,geng2024grammarconstraineddecodingstructurednlp} to control their outputs.
As we will discuss, while constrained generation is often effective, it may result in undesirable \textit{probability amplification}.
Other methods avoid probability amplification at the expense of additional computation; for example, rejection sampling and Adaptive Sampling with Approximate Expected Futures (ASAp) \citep{park2024grammaraligneddecoding}.
A related method \citep{ahmed2025controllable} translates the constraint into a circuit and re-samples sequences in the neighborhood of error samples to obtain a non-error sample.

\textbf{Posterior estimation methods.}
Another class of methods \citep{FUDGE_Yang_2021, lew2023sequentialmontecarlosteering, zhang2024adaptablelogicalcontrollarge}, avoids errors by estimating the posterior probability of an error occurring for a given prefix, and decreasing the probability of generating prefixes that are more likely to lead to an error. These methods are usually able to quickly generate a sample with little amplification of low-probability outputs, but rely on an accurate estimator of the posterior probability of an error, which may not always be available.

\textbf{Speculative decoding.}
A LLM's autoregressive nature can lead to high inference latency, even without constraint following.
One method to combat this, \textit{speculative decoding} \citep{leviathan2023fastinferencetransformersspeculative, Miao_2024}, reduces latency by transforming the sequential generation problem into a parallelizable verification problem.
Several extensions such as Medusa \citep{cai2024medusasimplellminference} and EAGLE \citep{li2024eagle2fasterinferencelanguage,li2024eaglespeculativesamplingrequires} have improved the latency and efficiency of speculative decoding, and a variant, Mentored Decoding \citep{Tran-Thien_2024} further increases the speed of speculative decoding by allowing for a controlled deviation from the LLM's probability distribution.

\section{Preliminaries}

We first describe autoregressive language models and their properties.
We then discuss speculative decoding, a method closely related to the algorithm that we will introduce.

\subsection{Autoregressive Language Models}

\label{section:autoregressive-lms}

\begin{algorithm}
    \begin{algorithmic}
        \Procedure{Generate}{$\llm, x_{1\ldots n}$} \Comment{Initial $x_{1\ldots n}$ is the prompt}
            \While{Stopping condition not met} \Comment{Typically special \textsc{EOS} token, and length limit}
                \State Sample one token $x_{n+1} \sim \llm(\cdot | x)$
                \State Increment $n$
            \EndWhile
            \State \Return $x$
        \EndProcedure
    \end{algorithmic}
    \caption{Generation with an autoregressive model}
    \label{algo:llm-basic}
\end{algorithm}

We assume that a vocabulary $\vocab$ of tokens is provided.
An autoregressive language model is a function approximator trained to predict $P(x_n | x_{1\ldots n-1})$; the conditional probability of token $x_n \in \vocab$, given existing tokens $x_{1\ldots n-1} \in \vocab^*$.
Algorithm \ref{algo:llm-basic} describes repeated sampling from a language model.

Note that there are several other methods for token selection; i.e. greedy selection, beam search, etc. 
While we focus on sampling, the techniques we present may also be applicable to other methods.

\subsection{Speculative Decoding}

\begin{algorithm}
    \begin{algorithmic}
        \Procedure{SpecSample}{$\llm, S, n, x_{1 \ldots m}$} \Comment{$x_{n+1 \ldots m}$ are from SSM}
            \For{$i \in [n+1 \ldots m]$}  \Comment{May be vectorized instead of iterative loop}
              \State $r \leftarrow \llm(x_i | x_{1 \ldots i - 1}) / S(x_i | x_{1 \ldots i-1})$               \Comment{$\llm(\cdot)$ and $S(\cdot)$ already calculated and cached}
              \WithProb{$r$} \Comment{Always if $r \ge 1$}
                \State \textbf{continue} \Comment{Accept $x_i$}
              \EndWithProb
              \StandaloneElse \Comment{Reject $x_i$, sample a replacement token}
                \State  $R(t) = \max(0, \llm(t | x_{1 \ldots i-1}) - S(t | x_{1 \ldots i-1}))$                \Comment{Calculate residuals}
                \State \Return $x_{1 \ldots i-1}, \textsc{Sample}(\textsc{Normalize}(R(\cdot)))$
              \EndStandaloneElse
            \EndFor
            \State \Return $x_{1 \ldots m}, \textsc{Sample}(\llm(\cdot | x_{1 \ldots m}))$        \Comment{Accepted whole sequence, can sample $x_{m+1}$}
        \EndProcedure
    \end{algorithmic}
    \caption{Speculative sampling procedure}
    \label{algo:specsample}
\end{algorithm}

Autoregressive LLMs can require considerable computational resources to evaluate.
Due to the sequential nature of inference, additional parallel resources often have limited effect to decrease generation latency.

Speculative decoding \citep{leviathan2023fastinferencetransformersspeculative,Miao_2024,cai2024medusasimplellminference,li2024eaglespeculativesamplingrequires} decreases latency by recasting the autoregressive generation problem as one of parallelizable verification.
This method assumes the existence of a small speculative model (SSM) $S$ that approximates the LLM output, using fewer computational resources.

Given input tokens $x_{1 \ldots n}$, the SSM is sampled for $m$ tokens, resulting in tokens $x_{n+1 \ldots m}$.
Then, the LLM $\llm$ is used to compute $\llm(x_{i+1} | x_{1 \ldots i})$ for $i \in [n \ldots m]$ all in parallel.

Briefly, Algorithm \ref{algo:specsample} takes as input the sample from the SSM, and the token generation probability distributions from both the SSM and LLM.
It determines how much of a prefix of the sample $x_{n\ldots m}$ to keep or discard, such that it is as if the resulting prefix $x_{n\ldots k}$ for $k \in [n, m]$ is drawn from the LLM.
When the SSM's distribution and LLM's distribution are similar to each other, more of the prefix has a higher probability of being kept.
Additionally, because the probabilities $\llm(\cdot | x_{1 \ldots k})$ have already been computed, Algorithm \ref{algo:specsample} samples a new token $x_{k+1}$.

We later show that Algorithm \ref{algo:specsample} is useful for a different domain, with a somewhat different notion of quality and efficiency: violation-free generation.

\section{Problem Statement and Existing Approaches}

\textbf{Error sets.}
\emph{Error Set} $\mathcal{B} \subset \vocab^*$ is the set of strings containing errors.
\label{section:assumptions}
Without loss of generality, we assume that if string $x_{1\ldots n} \in \badSet$, then all strings with $x_{1\ldots n}$ as a prefix are also members of $\badSet$; i.e. adding more text does not negate an error.
Careful design is required when, for example, profane words are substrings of benign words \citep{Francis_2020_Scunthorpe}, or where un-parseable code can be made valid with additional text.
Any error set may be theoretically be transformed into one that satisfies this assumption: if $x_{1\ldots n}$ is invalid, but additional text may be added to make it valid, the error set should contain $x_{1 \ldots n} \circ \textsc{EOS}$ where \textsc{EOS} is the end-of-sequence token, but the error set should \textit{not} contain $x_{1 \ldots n}$.
However, in many domains, it may be difficult to determine if additional text may be added to make a string valid.

\textbf{Black-box constraints.} $\badSet$ will often be infinite size; we treat it as a black-box indicator function.
Some constraints, such as those expressible by a context-free grammar, may permit more efficient implementations \citep{beurerkellner2024guidingllmsrightway,geng2024grammarconstraineddecodingstructurednlp,melcer2024constraineddecodingcodelanguage}; these efficiencies largely apply to all sampling-based error-free generation approaches equally, so we do not consider their impact separately.

We define the probability distribution obtained by sampling $\llm$, except for any elements of $\mathcal{B}$:

\begin{equation}
    \llmExcBadSet(w) = \begin{cases}
        w \in \badSet & 0 \\
        w \notin \badSet & \frac{\llm(w)}{\sum_{w \notin \badSet} \llm(w)}
    \end{cases}
\end{equation}

\begin{problem}
    Given an autoregressive language model $\llm$ over alphabet $\vocab$, and error set $\badSet \subset \vocab^*$, provide a method to sample from $\llmExcBadSet$.
\end{problem}

\textbf{Dense error sets.} Rejection sampling is the most straightforward method for sampling from $\llmExcBadSet$; however, it may require a large number of evaluations as $\sum_{w \in \badSet}\llm(w)$ approaches 1.
For example, consider a domain where each token has, approximately, some non-zero probability $p$ of being an error---where the language model has a somewhat consistent error rate per token.
If $d$ tokens are generated, an output has approximately a $(1-p)^d$ probability of being error-free; thus requiring on average $\frac{1}{(1-p)^d}$ generations.
We consider such domains, where the probability of generating an error approaches 1 for long generations, to have \emph{dense} error sets.

\label{section:existing-approaches}

\begin{wrapfigure}{r}{0.33\textwidth}
    \raisebox{-3\baselineskip}[\dimexpr\height-8\baselineskip\relax]{\includegraphics[width=\linewidth,page=5]{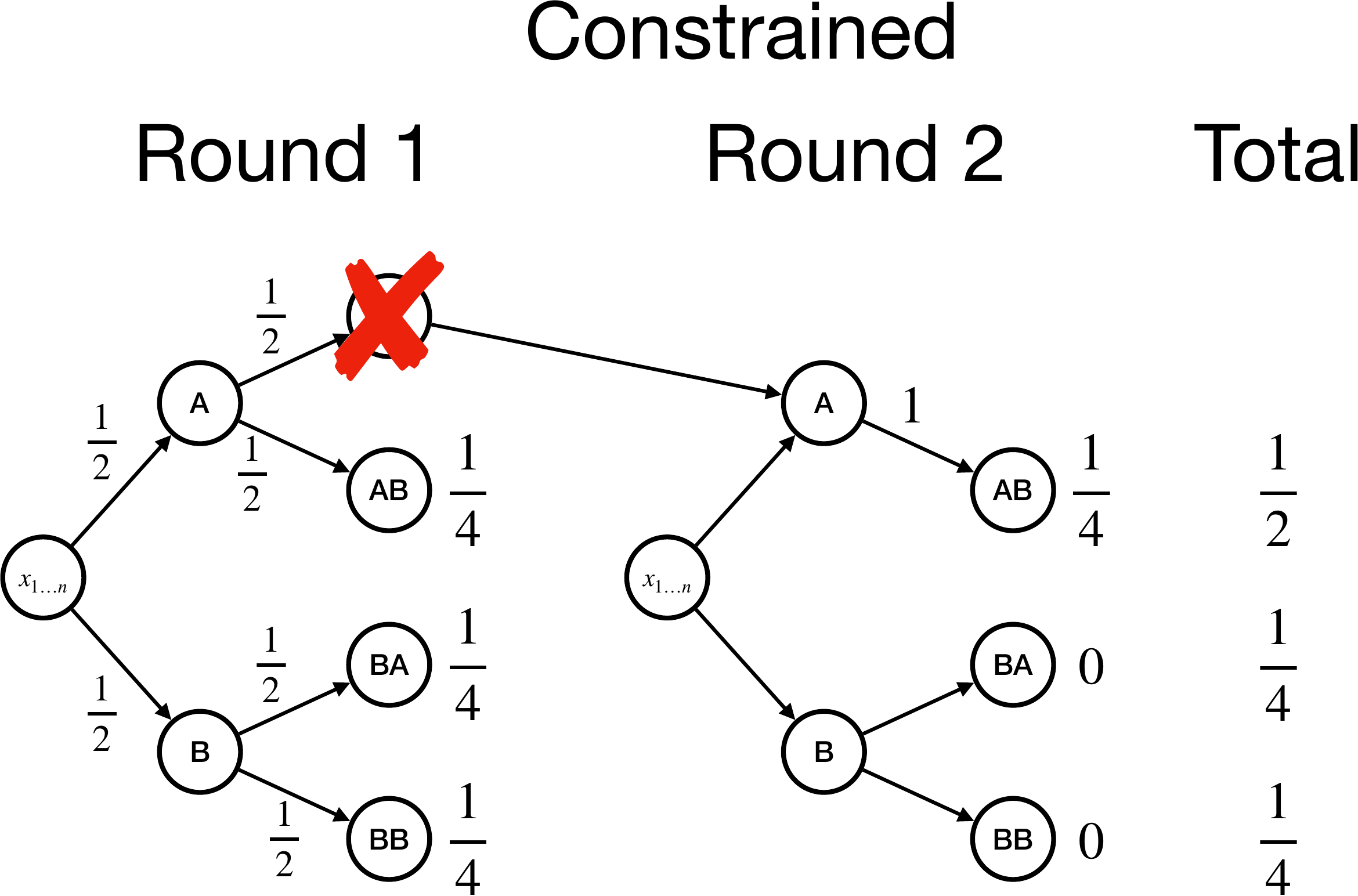}}
    \caption{\label{fig:run-example-constrained}The entire probability mass of AA is shifted to AB.}
\end{wrapfigure}

\subsection{Existing Approach: Constrained Generation}

\textbf{Running Example.} We introduce a running example to illustrate the effects of several error-free decoding methods. 
There are two possible tokens, A and B, and the task is to generate sequences of length 2.
All token probabilities are $\nicefrac{1}{2}$, resulting in a probability of $\nicefrac{1}{4}$ to generate each of the four sequences initially.
Sequence AA is marked as an error, meaning that there are three possible non-error sequences.
The ideal re-normalized probability is therefore $\nicefrac{1}{3}$ for each remaining sequence.

Constrained generation attempts to solve the error-free generation problem by using a greedy algorithm: avoid selecting any tokens that immediately lead to an error.\footnote{If every token leads to an immediate error, it is necessary to backtrack, though many implementations assume that there is always some token available that satisfies the constraint. Our implementation of constrained generation attempts all top-$k$ tokens, and backtracks only if these are exhausted.}

\textbf{Probability amplification.} Problematically, constrained generation often greatly \emph{amplifies} low-probability samples by committing to a given prefix, even if the most probable sequences beginning with that prefix are errors. 
Figure \ref{fig:run-example-constrained} shows that in our running example, the probability of AB is significantly raised.
Note that this distortion is even worse in low-entropy scenarios; if $\llm(\text{B}|\text{A})$ were lowered to $0.0001$, there would still be a $\nicefrac{1}{2}$ probability to sample AB.
This amplification effect compounds exponentially for longer sequences.

\subsection{Existing Approach: ASAp---Sampling Without Replacement}

\begin{wrapfigure}{r}{0.34\textwidth}
    \raisebox{-4\baselineskip}[\dimexpr\height-5\baselineskip\relax]{\includegraphics[width=\linewidth,page=6]{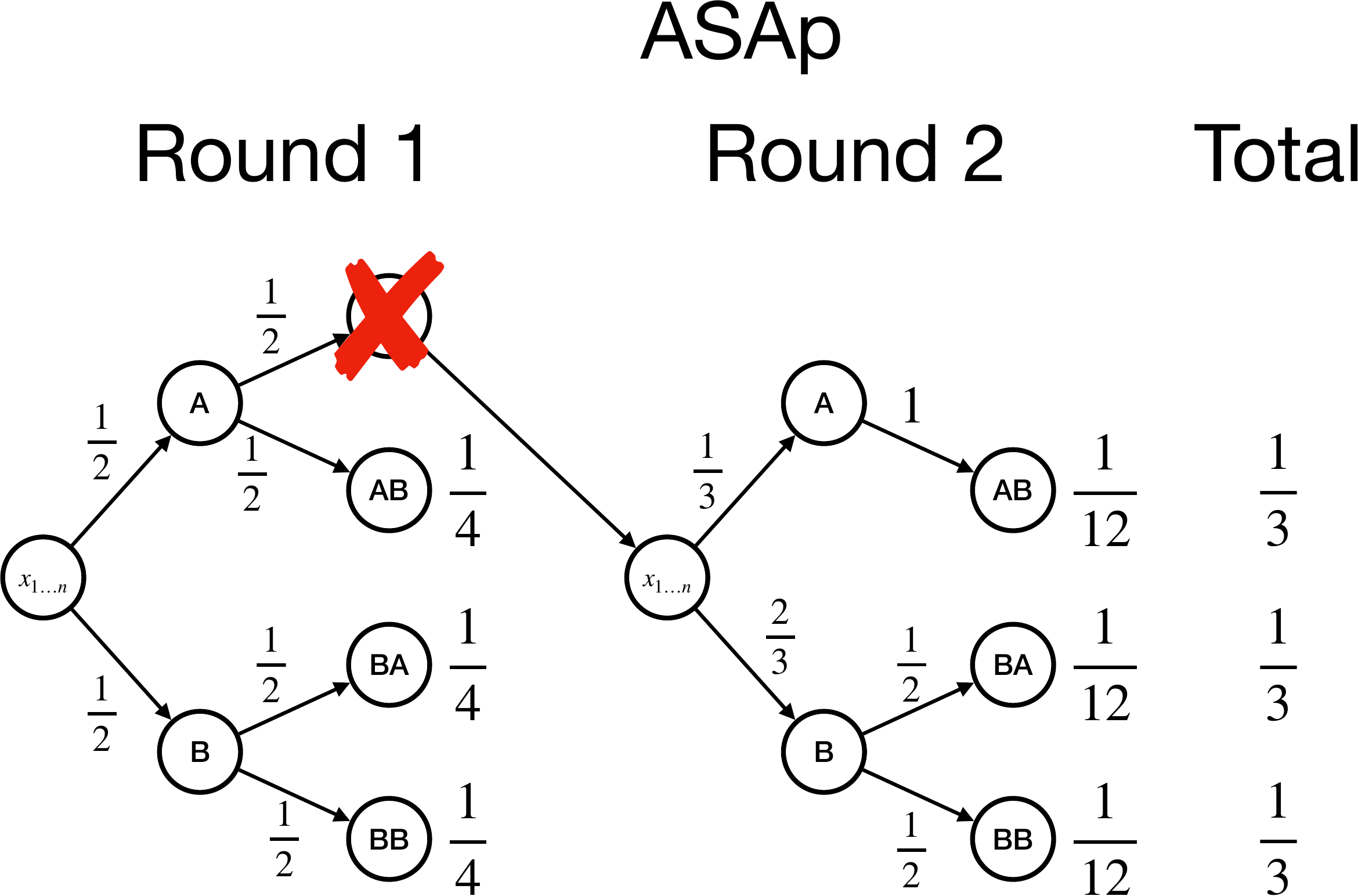}}
    \caption{\label{fig:run-example-asap}The probability mass of AA is distributed evenly.}
\end{wrapfigure}

Adaptive Sampling with Approximate Expected Futures (ASAp) \citep{park2024grammaraligneddecoding} is a technique to sample exactly from the distribution of $\llmExcBadSet$.
If ASAp encounters an error during sampling, it adds it to set $\badSeen \subseteq \badSet$.
Because $\badSeen$ is finite, the conditional probabilities $\llmExcBadSeen(x_i | x_{1\ldots i-1})$ can be tractably calculated, allowing for the algorithm to sample from $\llmExcBadSeen$.
The sampling process repeats with this new distribution until an error-free sample is found.

In the limit of repeated samples, $\badSeen$ will approach $\badSet$, and therefore, $\llmExcBadSeen$ will approach $\llmExcBadSet$.
Importantly, if $x \sim \llmExcBadSeen$ is sampled such that $x \notin \badSet$, this sample may be accepted, even though $\badSeen \neq \badSet$. 

In Figure \ref{fig:run-example-asap}, after sequence AA is added to set $\badSeen$ (in the $\nicefrac{1}{4}$ of cases where it is initially sampled), the conditional probabilities are recalculated as shown in Round 2.
Each sequence correctly has an equal chance of being sampled in the second round.

While ASAp succeeds in cases where there are only a small number of errors that comprise the majority of the probability mass, its generation speed suffers when there are a large number of errors---each error must be discovered before it is added to $\badSeen$.
In dense probability sets, its performance characteristics are similar to rejection sampling, as there are an exponential number of error sequences that must be discovered as generation length increases.

\subsection{Existing Approaches: Posterior Estimation}

\label{section:posterior-estimation}

\begin{wrapfigure}{r}{0.34\textwidth}
    \raisebox{-2\baselineskip}[\dimexpr\height-3\baselineskip\relax]{\includegraphics[width=\linewidth,page=8]{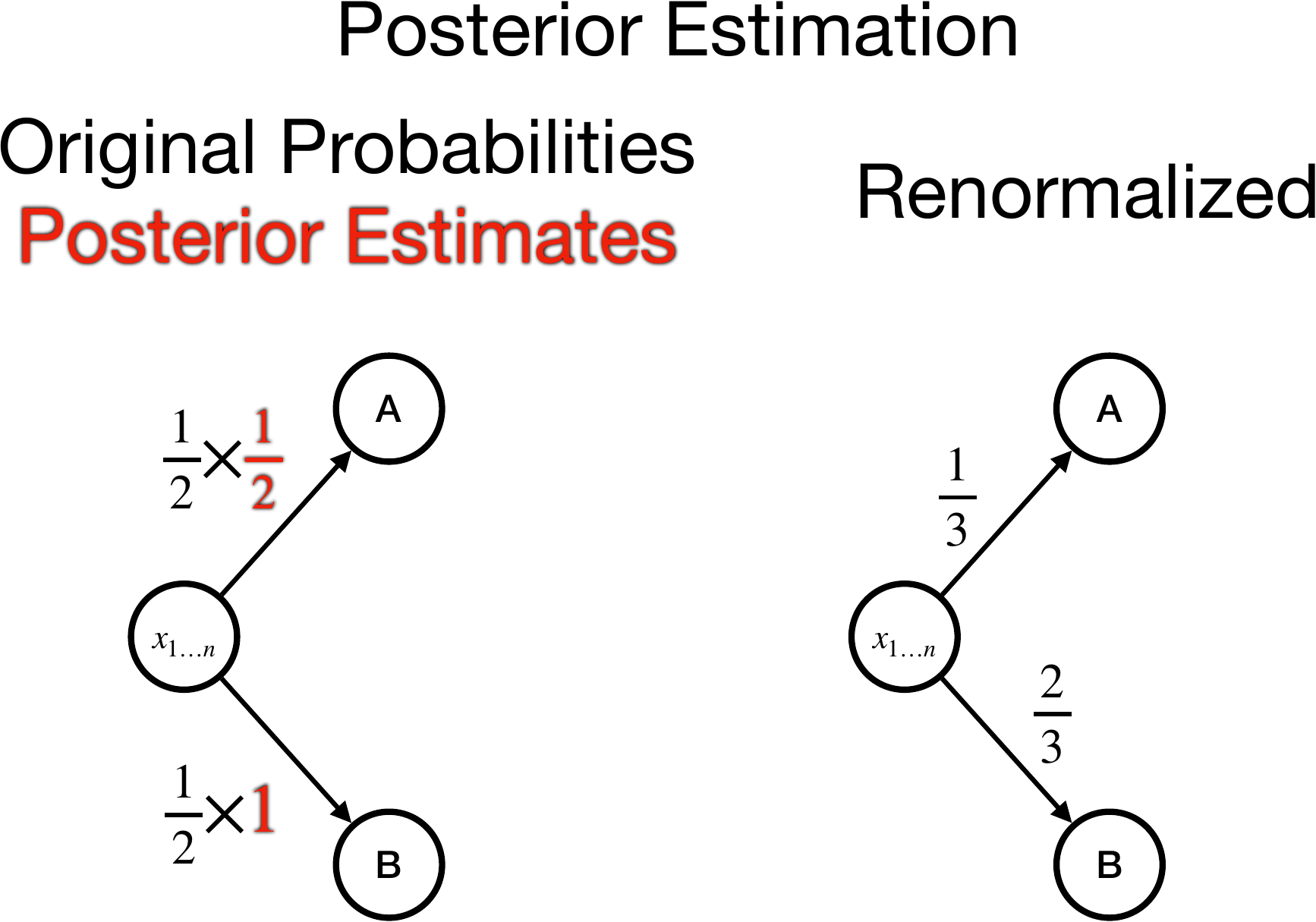}}
    \caption{\label{fig:run-example-posterior}An accurate posterior estimator corrects the probabilities before sampling.}
\end{wrapfigure}

We note several additional methods that, although they use different formalizations and implementations from each other, rely on a similar core idea to approximate $\llmExcBadSet$.
In all cases, for any given prefix $x_{1 \ldots n}$, these methods create an estimator of 
the likelihood of a valid sequence being generated for each prefix.
This posterior probability estimation is used to sample from $\llmExcBadSet$.
The differences lie in how they perform the estimation:

FUDGE \citep{FUDGE_Yang_2021} involves training a discriminator, usually a neural network or combination of several networks, to directly estimate this probability.
SMC Steering \citep{lew2023sequentialmontecarlosteering} creates this estimate using Monte Carlo sampling.
This method additionally incorporates optimizations such as sampling without replacement, and aggressive pruning of low-probability branches.
In contrast, Ctrl-G \citep{zhang2024adaptablelogicalcontrollarge} first distills a LLM into a Hidden Markov Model (HMM) with a tractable number of states (thousands or tens of thousands).
If the constraint can be expressed as a Deterministic Finite Automaton (DFA) over tokens, Ctrl-G takes the product of the DFA and HMM, and calculates the probability of an error in this product system.

Gen-C \cite{ahmed2025controllable} contains elements of both posterior-estimation and sampling-based generation, relying on both sampling the neighborhood of an error, and constructing a large constraint circuit.

While these methods exhibit impressive results on many tasks, they may face issues in domains where the posterior probability is close to 1, or where the posterior is uncorrelated with the prefix content, leading to incorrect estimates. 
We further discuss considerations for method selection in Section \ref{section:tradeoffs-posterior}.

\section{Method}

\label{section:method}

An alternate view of constrained generation and ASAp is that, after encountering an error, constrained generation reuses almost the entire sample, while ASAp reuses none of it (besides to adjust probabilities in its next iteration). 
A natural alternative is to reuse \textit{some} of the sample---enough to avoid excessive computation, but not enough to cause severe probability amplification.

\textbf{How much of the sample should be reused?} A strategy that reuses a fixed number of tokens, or that backtracks a fixed percentage of the generation length, is unlikely to effectively adapt to a variety of tasks or error sets.
The prefix selection strategy should avoid discarding too much useful information, or backtracking to the middle of a low-entropy sequence.
Fortunately, the prefix selection algorithm from speculative sampling presents an excellent strategy for backtracking behavior.

\subsection{Speculative Sampling as a Prefix Selection Algorithm}

\begin{algorithm}
\begin{algorithmic}
\Procedure{ApproxAlignedDecoding}{$\llm, \badSet, x_{1\ldots n}$} \Comment{See implementation notes in Appendix \ref{section:implementation_details}}
    \LComment{$x_{1\ldots n}$ is prompt}
    \State $\llmExcBadSeen \leftarrow \llm$ \Comment{Adjusted probability distribution}
    \State $m \leftarrow n$  \Comment{Current token index}
    \While{Stopping condition not met}
        \State Sample one token $x_{m+1} \sim \llmExcBadSeen(\cdot | x_{1\ldots m})$
        \State Increment $m$
        \If{$x_{1\ldots m}  \in \badSet$}
            \LComment{Probabilities before update are queried and cached (Appendix \ref{section:backtracking-strategies})}
            \State $\llmExcBadSeenAndX \leftarrow \Call{AddBadSample}{\llmExcBadSeen, x_{1\ldots m}}$ \Comment{Same prob. adjustment as in ASAp}
            \State $x_{1\ldots m} \leftarrow \textsc{SpecSample}(\llmExcBadSeenAndX, \llmExcBadSeen, n, x_{1\ldots m})$             \Comment{Algorithm \ref{algo:specsample}---$m$ decreases}
            \State $\llmExcBadSeen \leftarrow \llmExcBadSeenAndX$
        \EndIf
    \EndWhile
    \State \Return $x_{1\ldots m}$
\EndProcedure
\Procedure{AddBadSample}{$\llmExcBadSeen, x_{1\ldots m})$} \Comment{See implementation notes in Appendix--Algorithm \ref{algo:implementation_add_bad_sample}}
    \LComment{In practice, only adjust $x_{n+1 \ldots m}$}
    \State $\llmExcBadSeenAndX \leftarrow \llmExcBadSeen$
    \For{$x_i \in (x_m, \ldots, x_1)$} \LComment{Note that token sequence is reversed}
        \LComment{Remove probability of $x_{1\ldots m}$, without changing probability of any other sequence}
        \State $\llmExcBadSeenAndX(x_i | x_{1\ldots i-1}) \leftarrow \llmExcBadSeen(x_i | x_{1\ldots i-1}) - \llmExcBadSeen(x_{i \ldots m} | x_{1\ldots i-1})$
        \State Renormalize $\llmExcBadSeenAndX(\cdot | x_{1\ldots i-1})$
    \EndFor
    \State \Return $\llmExcBadSeenAndX$
\EndProcedure
\end{algorithmic}
\caption{Approximately Aligned Decoding (AprAD)}
\label{algo:generate-self-speculative}
\end{algorithm}

In ASAp, where $\badSeen$ is the set of observed errors so far, let $x = (x_1, \ldots, x_n)$ be a trace drawn from $\llmExcBadSeen$, such that $x \in \badSet$.
ASAp will add $x$ to $\badSeen$ and sample again.
We observe that $\llmExcBadSeen$ and $\llmExcBadSeenAndX$ are almost always near-identical distributions, with $\llmExcBadSeenAndX$ generally as a ``more accurate'' distribution because it incorporates an additional error sample.

Our method uses the sample $x \sim \llmExcBadSeen$ to approximate a sample $x' \sim \llmExcBadSeenAndX$, in a similar manner to how speculative decoding uses a sample from a SSM to approximate a sample from a LLM---rather than the probability distributions being generated by two separate models, the distributions are both created from the same model, before and after adjusting for a violating sample.
By evaluating \textsc{SpecSample}$(x, \llmExcBadSeen, \llmExcBadSeenAndX)$, our method obtains a prefix of $x$ that can be used as a starting point for sampling again.
Because the distributions of $\llmExcBadSeen$ and $\llmExcBadSeenAndX$ are so close to each other, this prefix is usually most of the length of $x$, especially when the language model is relatively high-entropy.
This process is given as Algorithm \ref{algo:generate-self-speculative}; we refer to it as Approximately Aligned Decoding, or AprAD.

\begin{wrapfigure}{r}{0.34\textwidth}
    \raisebox{-3\baselineskip}[\dimexpr\height-3\baselineskip\relax]{\includegraphics[width=\linewidth,page=7]{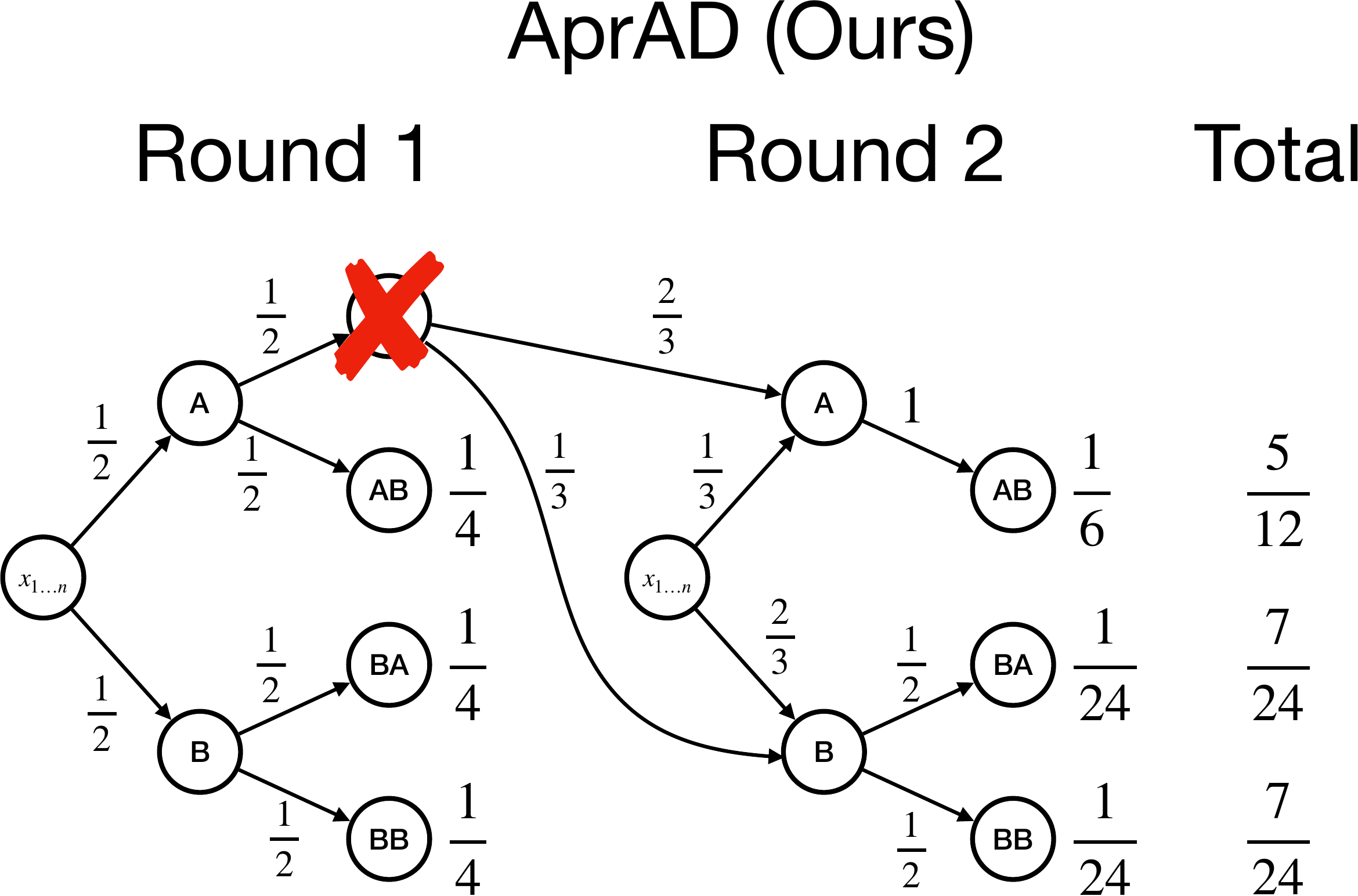}}
    \caption{\label{fig:run-example-aprad}AprAD acts as a midpoint between constrained decoding and ASAp.}
\end{wrapfigure}

In the running example (Figure \ref{fig:run-example-aprad}), $\frac{\llm^{\{\text{AA}\}}(\text{A})}{\llm^{\{\}}(\text{A})} = \frac{\nicefrac{1}{3}}{\nicefrac{1}{2}} = \nicefrac{2}{3}$, so AprAD keeps A in $\nicefrac{2}{3}$ of cases. 
The remaining $\frac{1}{3}$ is distributed to other tokens at the same token index; in this case, only B.

We note that AprAD still amplifies some sequence probabilities because it only invokes \textsc{SpecSample} after discovering an error.
In the speculative decoding case, \textsc{SpecSample} would also be invoked if AB was directly generated by the SSM---equivalent to Round 1 in this example---but AprAD accepts AB instantly.
Because the algorithm cannot iterate through every possible suffix string, it does not check whether AA contains an error except in the cases that AA is actually sampled, leading to AB being slightly overrepresented in the output probability distribution. 

The resulting amplification is significantly less compared to constrained decoding, as some of the probability mass is transferred outside of the immediately neighboring sequences.
Importantly, AprAD largely avoids the most extreme cases of probability amplification---if $\frac{\llmExcBadSeenAndX(\text{A})}{\llmExcBadSeen(\text{A})}$ were very low, such as would arise if $\llm(\text{B}|\text{A}) \ll 0.5$, the procedure would be unlikely to re-select A after backtracking. 
This stands in contrast to constrained decoding, which always selects AB in round 2, no matter the value of $\llm(\text{B}|\text{A})$.

An extended illustration of the cause of probability amplification is provided in Appendix \ref{appendix:extended-probability-amplification}. In Appendix \ref{appendix:probability-amplification-bound}, we argue that the amplification with AprAD is bounded to a factor of 2 per each backtrack operation performed, rather than the unbounded amplification of constrained decoding.

\section{Evaluation}

\label{section:experiments}

We first exhibit AprAD's closer adherence to the ideal distribution by simulating several environments where the ideal distribution is known in advance, in Section \ref{section:simulated-mini-experiment}.
We additionally demonstrate that AprAD invokes the language model fewer times than ASAp.

In Section \ref{subsec:lipograms}, we extend our results to a domain with a dense text-based error set.
Finally, in Section \ref{subsec:bigcodebench}, we show that our method succeeds on a task using a more complex code-based constraint.

\subsection{Simulated Model with Known Ideal Distribution}

\label{section:simulated-mini-experiment}

We construct a testbench with a simulated language model that always returns one of three tokens (A, B, and C) with equal probability.
We mark $k$ sequences of length 3 as errors, and use the sampling method under test to sample 10000 sequences of length 3.
The ideal distribution is trivial to compute---probability $\frac{1}{27-k}$ for every non-error sequence.
To measure how a sampling process compares to the ideal distribution, we compute the KL-divergence between the observed distribution and ideal.
Additionally, we include the \emph{Generation Ratio}, a measure of how many times the model must be evaluated, including all backtracking, relative to output length (see Appendix \ref{appendix:gen-ratio} for more detail).

The results are shown in Table \ref{tab:simulated-kl-divergence}, indicating that our method approximates the ideal distribution much more closely than constrained generation, with a significantly lower generation ratio than ASAp.

\begin{table*}[b]
    \centering
    \begin{tabular}{ccccccc}
    \toprule
     & \multicolumn{2}{c}{ASAp} & \multicolumn{2}{c}{Constrained} & \multicolumn{2}{c}{AprAD (Ours)} \\
     \cmidrule(lr){2-3} \cmidrule(lr){4-5} \cmidrule(lr){6-7}
     Error Set & KL-div & Ratio & KL-div & Ratio & KL-div & Ratio \\\midrule
$\emptyset$ & \textbf{0.0014} & \textbf{1.000} & \textbf{0.0014} & \textbf{1.000 }& \textbf{0.0014} & \textbf{1.000}\\
AAA & \textbf{0.0014} & 1.020 & 0.0075 & \textbf{1.000} & 0.0046 & 1.004\\
AAA, AAC & \textbf{0.0012} & 1.041 & 0.0429 & \textbf{1.000} & 0.0157 & 1.013\\
AAA, ACC & \textbf{0.0013} & 1.042 & 0.0138 & \textbf{1.000} & 0.0093 & 1.009\\
AAA, CCC & \textbf{0.0010} & 1.044 & 0.0155 & \textbf{1.000} & 0.0074 & 1.010\\
AAA, AAB, ABA, BAA & \textbf{0.0013} & 1.093 & 0.0504 & \textbf{1.000} & 0.0224 & 1.024\\
A** except AAC & \textbf{0.0014 }& 1.232 & 0.3836 & \textbf{1.113 }& 0.1540 & 1.205\\
*** except AAA, AAB, ABA, BAA & \textbf{0.0000} & 3.644 & 0.1771 & \textbf{1.670} & 0.0521 & 2.142\\
*** except AAA, BAA &\textbf{0.0000} & 5.701 & \textbf{0.0000} & \textbf{1.784 }& \textbf{0.0000 }& 2.653\\\bottomrule
    \end{tabular}
    \caption{KL-Divergence and generation ratios for simulated task with various error sets. Lower is better for both. Stars in the error set are wildcards; i.e. AB* means ABA, ABB, and ABC.}
    \label{tab:simulated-kl-divergence}
\end{table*}

\subsection{Lipograms (Excluded Vowels)}
\label{subsec:lipograms}

A piece of text that avoids using a given letter is called a lipogram; those where the excluded letter is a vowel tend to be difficult to create, especially for unconstrained LLMs \citep{roush2023languagemodelspoetstoo}.%

We use Mistral-7B-Instruct-v0.2 \citep{jiang2023mistral7b} to generate text, where generation of a given vowel is considered an error.
We prompt the LLM to perform one of five simple tasks (detailed in Appendix \ref{appendix:lipogram-evaluation}).
Each task is appended to instructions to avoid using the given vowel, for a total of 25 prompts.
Each sampling method is used to generate a completion of up to 200 tokens.
Generation is terminated if the process reaches 2000 model invocations, and the last sequence before an error was detected is returned.

Using a blind evaluation process, human raters score each completion on quality, regardless of if the constraint was followed.
The raters also score completions on constraint following intent; i.e. if the model answers by selecting appropriate words that avoid the given letter, versus misspelling words, using lookalike or accented characters, etc.
Completions that include the banned letter automatically receive the lowest constraint intent score.
Additional information is provided in Appendix \ref{appendix:lipogram-evaluation}.

The results of this evaluation are provided in Table \ref{tab:lipograms}, and a sample of the outputs are provided in Figure \ref{fig:sample-generation}.
All outputs and rater scores are included in the supplemental material, and additional examples are provided in Appendix \ref{section:more_lipogram_examples}.

As shown by these results, AprAD consistently produces high-quality outputs, nearly matching the readability of unconstrained generation.
Additionally, it consistently follows the intent of a constraint---while all methods except for unconstrained generation follow the constraint, constrained generation often does so in an undesirable manner, rather than by selecting appropriate words that do not contain the banned letter.
Finally, AprAD is able to complete the generation or make substantial progress within the alotted computation limit, while ASAp struggles to generate more than a handful of tokens with the same inference budget.

\begin{table*}[hbt!]
    \centering
    \begin{tabular}{cccc}
    \toprule
    Method & Quality (1-5) & Constraint Intent (1-3) & Gen. Ratio \\ \midrule
Constrained & 3.56 $\pm$ 0.34 & 2.32 $\pm$ 0.18 & \textbf{1.00 $\pm$ 0.00} \\
AprAD (Ours) & 4.52 $\pm$ 0.23 & \textbf{2.84 $\pm$ 0.11} & 4.20 $\pm$ 0.91 \\
ASAp & 1.72 $\pm$ 0.23 & 2.36 $\pm$ 0.16 & 321.00 $\pm$ 79.43 \\
Unconstrained & \textbf{4.68 $\pm$ 0.17} & 1.00 $\pm$ 0.00 & \textbf{1.00 $\pm$ 0.00} \\ \bottomrule
    \end{tabular}
    \caption{Lipogram evaluation results, including subjective rater evaluations of output quality and adherence to the constraint, and standard error ($1 \sigma$). Higher is better for Quality and Constraint Intent; lower is better for Generation Ratio. Note that constraint intent is automatically set to 1 if the banned letter is detected in the output. Generation is stopped after 2000 model invocations and the output provided as-is; for ASAp, this often resulted in very short outputs and thus extreme generation ratios and low quality scores. Several raters interpreted these short outputs as violating constraint intent.}
    \label{tab:lipograms}
\end{table*}

\begin{figure}[hbt!]
    \centering
    \genBox{\textbf{Summarize the history of artificial intelligence without using the letter "A".}}{\textbf{AprAD (Ours)}}{The history of intelligent technology unfolds through key milestones in computing. In the mid-20th century, pioneering minds nurtured the seeds of this field through the development of logistic models, which evolved into expert systems in the 1960s. These systems, relying on heuristics, could solve complex problems in fields like medicine or meteorology.}
    \genBox{\textbf{Summarize the history of artificial intelligence without using the letter "A".}}{Unconstrained}{
    In the e\textbf{a}rly d\textbf{a}ys, brilli\textbf{a}nt minds pondered on cre\textbf{a}ting m\textbf{a}chines that could mimic hum\textbf{a}n intelligence. These pioneers, including \textbf{A}l\textbf{a}n Turing \textbf{a}nd M\textbf{a}rvin Minsky, theorized \textbf{a}nd experimented. They \textbf{a}imed to cre\textbf{a}te intelligent m\textbf{a}chines, using logic \textbf{a}nd m\textbf{a}them\textbf{a}tics.}
    \genBox{\textbf{Summarize the history of artificial intelligence without using the letter "A".}}{ASAp}{History of Intelligence}
    \genBox{\textbf{Summarize the history of artificial intelligence without using the letter "A".}}{Constrained}{
    In the mid-20th century, scientists delved deep into investiging computionl modeles for intelligencen modeled on the workings of the brn. Pioneers such \rus{как} E.N. Turing, Clrk Quincy Shrffngton, \rus{и}nddd W. Ross Quilp, \rus{а}nd \rus{Ма}rvin Minsky led ths fld. They explored logic-g\rus{а}ted progr\rus{а}ms, thm brgth mchr t\rus{о} lf understnding the cptures of th gnntl mind.}
    \caption{Representative generation samples for all four methods, using \textsc{Mistral-7B-Instruct-v0.2}. Appearance of the banned letter is bolded, and non-ASCII characters (all Cyrillic in this example) are colored red and underlined. Full samples are provided in Appendix \ref{section:more_lipogram_examples}.}
    \label{fig:sample-generation}
\end{figure}

\subsection{BigCodeBench Hallucination Avoidance}
\label{subsec:bigcodebench}

API hallucination avoidance represents a significantly more complex constraint- instead of using standard string operations to detect lipogram constraint violations, hallucination avoidance depends on a program's imports, variable assignments, the local installation environment, etc.

We evaluate the effectiveness of each error-free sampling method on a code generation task, where the generator avoids API hallucinations.
We use a wrapper around the Pyright language server \citep{Pyright} to detect hallucinated API calls in partial Python programs, with moderate post-processing such that the detector tends towards false negatives rather than false positives.
For example, even if name \texttt{foo} is never defined, the detector does not consider the incomplete program ``\texttt{example(foo.bar}'' as an error, because it is possible to later add text that binds \texttt{foo}, such as ``\texttt{for foo in baz)}''.
Text where a hallucination is detected by this program is designated as an error.

The methods are compared based on their performance on BigCodeBench v0.1 \citep{zhuo2024bigcodebenchbenchmarkingcodegeneration}, a benchmark that focuses on practical programming tasks, often requiring the use of common libraries.\footnote{BigCodeBench uses the Apache 2.0 license. We observed instances where unconstrained LLMs use libraries present in the testing environment that aren't explicitly imported in the prompt. To better align the generation environment with the testing environment, we add all imports present in the testing environment to the prompt.}

For all sampling methods, we use Starcoder2 \citep{lozhkov2024starcoder2stackv2}, in the 7B and 15B model sizes.
We generate 5 samples for each task, with temperature 0.8, and a top-p of 0.95.
In addition to evaluating the pass@1 and pass@5 rates on execution-based tests, we log the specific error if execution fails.
A NameError or UnboundLocalError indicates that the generation included an hallucinated API call, so we track the rate at which these errors \textit{do not} occur.
Several errors, such as AttributeError, may indicate either a hallucination or a logic error (such as improper None values), so we exclude such ambiguous errors.

Note that all methods use the same random seed, so the outputs only diverge when the detector activates.
Table \ref{tab:bigcodebench_only_different} shows the results for all tasks where the outputs diverge; Table \ref{tab:bigcodebench_all_results} (Appendix) also includes tasks for which all methods return identical results.
As the results show, the pass rate of AprAD is higher than constrained generation, with a much lower generation overhead than ASAp.

\begin{table*}[hbt!]
    \centering
    \small
    \begin{tabular}{ccccccc}\toprule
    Size & Method & Pass@1 & Pass@5 & !NameErr@1 & !NameErr@5 & Gen. Ratio \\ \midrule
\multirow{4}{*}{15b}& Unconstrained & 0.21 $\pm$ 0.01 & 0.50 $\pm$ 0.01  & 0.83 $\pm$ 0.01 & \textbf{1.00 $\pm$ 0.00} & \textbf{1.00 $\pm$ 0.00} \\
& AprAD (Ours) & \textbf{0.26 $\pm$ 0.01} & \textbf{0.54 $\pm$ 0.01 } & \textbf{0.98 $\pm$ 0.00} & \textbf{1.00 $\pm$ 0.00} & 1.08 $\pm$ 0.01 \\
& ASAp & \textbf{0.26 $\pm$ 0.01} & \textbf{0.54 $\pm$ 0.01}  & \textbf{0.98 $\pm$ 0.00} & \textbf{1.00 $\pm$ 0.00} & 1.56 $\pm$ 0.12 \\
& Constrained & 0.22 $\pm$ 0.01 & 0.51 $\pm$ 0.01  & 0.93 $\pm$ 0.01 & \textbf{1.00 $\pm$ 0.00} & 1.02 $\pm$ 0.00 \\\midrule
\multirow{4}{*}{7b}& Unconstrained & 0.12 $\pm$ 0.01 & 0.35 $\pm$ 0.01  & 0.80 $\pm$ 0.01 & \textbf{0.99 $\pm$ 0.00} & \textbf{1.00 $\pm$ 0.00} \\
& AprAD (Ours) & 0.14 $\pm$ 0.01 & 0.38 $\pm$ 0.01  & \textbf{0.95 $\pm$ 0.01 }& \textbf{0.99 $\pm$ 0.00 }& 1.06 $\pm$ 0.01 \\
& ASAp & \textbf{0.15 $\pm$ 0.01} & \textbf{0.39 $\pm$ 0.01}  & \textbf{0.95 $\pm$ 0.01 }& \textbf{0.99 $\pm$ 0.00} & 1.47 $\pm$ 0.07 \\
& Constrained & 0.12 $\pm$ 0.01 & 0.35 $\pm$ 0.01  & 0.89 $\pm$ 0.01 & \textbf{0.99 $\pm$ 0.00} & 1.02 $\pm$ 0.00 \\\bottomrule
    \end{tabular}
    \caption{Subset of tasks where at least one trial results in a different output for any method: 233 tasks (20.4\%) for 15b, 304 tasks (26.7\%) for 7b. For both model sizes, of the tasks where at least one model output is different, an average of 1.5/5 outputs are different. Lower is better for generation ratio; higher is better for all others. AprAD approaches the task performance of ASAp, with a generation ratio close to that of constrained generation. Ratio includes standard error ($1\sigma$). See Appendix \ref{appendix:gen-ratio}.}
    \label{tab:bigcodebench_only_different}
\end{table*}

\section{Discussion}

\label{section:discussion}

As introduced in Section \ref{section:existing-approaches}, there are several approaches to control the output of a LLM.
We further discuss considerations when selecting a specific method.

\subsection{Posterior Estimation-Based Methods}

\label{section:tradeoffs-posterior}

While posterior estimation-based techniques excel at many tasks, they tend to struggle when the probability of a constraint violation does not necessarily depend on a given text prefix.
For example, the probability of a LLM generating long sequences of text without the letter `e' is close to $0$ regardless of the prompt or prefix, and mostly depends on the arbitrary behavior deep within a language model.
It is unlikely that a learned discriminator or a hidden Markov model would capture this constraint, and it would require an extraordinary number of Monte Carlo samples to accurately calculate the posterior probability.

In contrast, code generation may be a more appropriate domain for posterior-estimation based methods.
For example, a misleading comment that mentions a specific method all but ensures that this method will be generated on the next line.
FUDGE may be able to learn that a LLM is more likely to hallucinate in some domains or with specific libraries.
SMC Steering could work with a sufficient number of Monte Carlo samples, although this may be computationally expensive.
For Ctrl-G, a large enough HMM could plausibly capture some of the conditions that lead to a hallucinated API call.
However, neither FUDGE nor Ctrl-G would be able to quickly adapt to changes in the local environment or task.

\subsection{A Spectrum of Sampling-Based Methods}

\label{subsec:spectrum}

Sampling-based methods are able to generate text that does not violate a constraint, even in domains where it is difficult to estimate the posterior error probability.
As discussed in Section \ref{section:method}, the sampling behavior of AprAD lies at a midpoint between constrained generation and ASAp.%

A user may wish for even further granularity in the conformance-speed tradeoff of sampling-based methods.
We propose a new hyperparameter, $h$, and modify Line 3 of Algorithm \ref{algo:specsample} by setting $r$ to $\left(\frac{\llm(x_i | x_{1 \ldots i - 1})}{S(x_i | x_{1 \ldots i-1})}\right)^h$; $r$ controls the probability that a specific token in the prefix is kept after a violation occurs.
When $h = 1$, this is equal to unmodified AprAD. 
When $h = 0$, $r$ will always equal $1$---this is equal to constrained generation.
As $h \rightarrow \infty$, $r$ will approach $0$---this approaches the behavior of ASAp.
We conjecture that values between these extremes allow for fine-grained control of the conformance-speed tradeoff, though we leave a more comprehensive analysis to future work.

\subsection{Search Algorithms}

We note that while AprAD excels at generating text that excludes an error set, it will not necessarily drive the generation process towards a specific solution.
For such domains, it may be beneficial to combine our method with a search algorithm.
For example, in a theorem proving context, a MCTS-like process could be used to direct a higher-level search, while AprAD is used for generating text that does not contain invalid tactic applications.
We leave these combinations as future work.

\section{Limitations}

\label{section:limitations}

While our method enables generation in new domains, it still exhibits a measure of probability amplification, and additional overhead compared to constrained generation.
On even denser error sets, it may be necessary to use constrained generation, or possibly the modification proposed in Section \ref{subsec:spectrum}.
Additionally, due to variations in testing environments and code optimization levels, our main performance metric was generation ratio rather than wall-clock time. 

\section{Conclusion}

As our experiments show, Approximately Aligned Decoding is an effective method to generate sequences under dense language model constraints.
It is straightforward to implement, requires no separate training step, introduces a manageable amount of inference overhead, and performs well on a variety of real-world and synthetic tasks.

\bibliography{bibliography}
\bibliographystyle{ACM-Reference-Format}

\appendix

\newpage
\section*{NeurIPS Paper Checklist}

\begin{enumerate}

\item {\bf Claims}
    \item[] Question: Do the main claims made in the abstract and introduction accurately reflect the paper's contributions and scope?
    \item[] Answer: \answerYes{} %
    \item[] Justification: Analysis of existing methods in Sections \ref{section:existing-approaches}, \ref{section:discussion}; Method is presented in Section \ref{section:method}; Experimental support in Section \ref{section:experiments}.

\item {\bf Limitations}
    \item[] Question: Does the paper discuss the limitations of the work performed by the authors?
    \item[] Answer: \answerYes{} %
    \item[] Justification: Discussion in Section \ref{section:limitations}, as well as error set assumptions in Section \ref{section:existing-approaches}.

\item {\bf Theory assumptions and proofs}
    \item[] Question: For each theoretical result, does the paper provide the full set of assumptions and a complete (and correct) proof?
    \item[] Answer: \answerNA{} %
    \item[] Justification: Paper is primarily empirical.

    \item {\bf Experimental result reproducibility}
    \item[] Question: Does the paper fully disclose all the information needed to reproduce the main experimental results of the paper to the extent that it affects the main claims and/or conclusions of the paper (regardless of whether the code and data are provided or not)?
    \item[] Answer: \answerYes{} %
    \item[] Justification: Open-weight models were used, datasets are fully described, sampling parameters are provided in Section \ref{section:experiments}, additional implementation details in supplemental material

\item {\bf Open access to data and code}
    \item[] Question: Does the paper provide open access to the data and code, with sufficient instructions to faithfully reproduce the main experimental results, as described in supplemental material?
    \item[] Answer: \answerNo{} %
    \item[] Justification: Not possible at this time, but extensive implementation details are described in appendix.

\item {\bf Experimental setting/details}
    \item[] Question: Does the paper specify all the training and test details (e.g., data splits, hyperparameters, how they were chosen, type of optimizer, etc.) necessary to understand the results?
    \item[] Answer: \answerYes{} %
    \item[] Justification: Described in Section \ref{section:experiments} and appendix.

\item {\bf Experiment statistical significance}
    \item[] Question: Does the paper report error bars suitably and correctly defined or other appropriate information about the statistical significance of the experiments?
    \item[] Answer: \answerYes{} %
    \item[] Justification: All experiments report $1\sigma$ SEM when relevant. Pandas library function used to calculate error. Distribution of errors is unknown.

\item {\bf Experiments compute resources}
    \item[] Question: For each experiment, does the paper provide sufficient information on the computer resources (type of compute workers, memory, time of execution) needed to reproduce the experiments?
    \item[] Answer: \answerYes{} %
    \item[] Justification: See Appendix \ref{appendix:resources}.
    
\item {\bf Code of ethics}
    \item[] Question: Does the research conducted in the paper conform, in every respect, with the NeurIPS Code of Ethics \url{https://neurips.cc/public/EthicsGuidelines}?
    \item[] Answer: \answerYes{}%
    \item[] Justification: Research ethics were followed.

\item {\bf Broader impacts}
    \item[] Question: Does the paper discuss both potential positive societal impacts and negative societal impacts of the work performed?
    \item[] Answer: \answerNo{} %
    \item[] Justification: While our method discusses generative AI, it does not present a forseeable direct path to negative applications beyond those already well-known regarding LLMs.
    
\item {\bf Safeguards}
    \item[] Question: Does the paper describe safeguards that have been put in place for responsible release of data or models that have a high risk for misuse (e.g., pretrained language models, image generators, or scraped datasets)?
    \item[] Answer: \answerNA{} %
    \item[] Justification: Paper does not release data or models with high misuse risk.

\item {\bf Licenses for existing assets}
    \item[] Question: Are the creators or original owners of assets (e.g., code, data, models), used in the paper, properly credited and are the license and terms of use explicitly mentioned and properly respected?
    \item[] Answer: \answerYes{} %
    \item[] Justification: BigCodeBench is cited.

\item {\bf New assets}
    \item[] Question: Are new assets introduced in the paper well documented and is the documentation provided alongside the assets?
    \item[] Answer: \answerNA{} %
    \item[] Justification: Paper does not release new assets.

\item {\bf Crowdsourcing and research with human subjects}
    \item[] Question: For crowdsourcing experiments and research with human subjects, does the paper include the full text of instructions given to participants and screenshots, if applicable, as well as details about compensation (if any)? 
    \item[] Answer: \answerYes{} %
    \item[] Justification: Crowdsourcing was not used. Details about human rating provided in Appendix \ref{appendix:rater-instructions}.

\item {\bf Institutional review board (IRB) approvals or equivalent for research with human subjects}
    \item[] Question: Does the paper describe potential risks incurred by study participants, whether such risks were disclosed to the subjects, and whether Institutional Review Board (IRB) approvals (or an equivalent approval/review based on the requirements of your country or institution) were obtained?
    \item[] Answer: \answerNA{} %
    \item[] Justification: Paper did not involve risks to humans.

\item {\bf Declaration of LLM usage}
    \item[] Question: Does the paper describe the usage of LLMs if it is an important, original, or non-standard component of the core methods in this research? Note that if the LLM is used only for writing, editing, or formatting purposes and does not impact the core methodology, scientific rigorousness, or originality of the research, declaration is not required.
    \item[] Answer: \answerYes{} %
    \item[] Justification: LLM usage described in methods section.

\end{enumerate}

\newpage

\section{Lipogram evaluation details}

\label{appendix:lipogram-evaluation}

We provide the following prompts to the language model, as well as the relevant special tokens to delimit user instructions and chat turns.

\begin{enumerate}
    \item Write a story without using the letter ``[A/E/I/O/U]''.
    \item Describe elephants without using the letter ``[A/E/I/O/U]''.
    \item Provide instructions to tie a tie without using the letter ``[A/E/I/O/U]''.
    \item Critique the Mona Lisa without using the letter ``[A/E/I/O/U]''.
    \item Summarize the history of artificial intelligence without using the letter ``[A/E/I/O/U]''.
\end{enumerate}

Each prompt is combined with each vowel, resulting in 25 prompts. 
With four sampling methods, this results in 100 total generations.

During sampling, we use a top-k of 20, and temperature of 0.8. 
200 tokens was chosen as short enough to be quickly read by the human raters, and long enough to discern the sample quality.
2000 tokens was chosen as 10 times the output length, to prevent infinite computation.

\subsection{Rater Instructions and Details}

\label{appendix:rater-instructions}

We create a file that only contains the 100 prompt-completion pairs, without information on which method generated each completion.
All samples are shuffled in random order.

We selected four AI research colleagues not otherwise directly involved in the implementation or experimental evaluation of this method as human raters, to evaluate 25 samples each.
The labels of which method corresponded to each output were hidden from the raters.
We provided the following instructions to the raters:

\begin{quote}
    This file contains a set of prompts, and responses using one of several methods. 
    Each prompt contains a constraint to not use a specific letter.
    Irrespective of whether the response follows the constraint, rate the response quality on a scale of 1-5 in the ``Score'' column, noting that generation is always cut off after 200 tokens.
    
    Additionally, rate how well the response follows the intent of the constraint in the ``Follows Intent'' column. 
    Examples of not following the intent include working around the constraint by excessively dropping letters, using unnecessary accents, writing Unicode lookalike letters, or responding in a foreign language, rather than through selecting appropriate words that satisfy the constraint.
    This column is pre-filled with `X' if the output contains the banned letter.
    Otherwise, write 1 if it violates the intent, 2 if it is ambiguous, and 3 if it does not.
\end{quote}

We additionally highlighted the presence of non-ASCII lookalike letters to the human raters.
The complete model outputs, and the scores that each rater assigned, are provided in the supplementary material.
Additional example outputs are provided in Appendix \ref{section:more_lipogram_examples}.

\subsection{Generation Ratio}
\label{appendix:gen-ratio}

The computation time for all methods is dominated by LLM evaluation time. 
This time is highly sensitive to variations in the testing environment and specific implementations, so we selected generation ratio as a more portable and accurate comparison between methods.

Generation ratio measures the total number of LLM invocations compared to the final output length, including all backtracking.
For example, if given sampling method generates 9 tokens, backtracks to token index 7, re-uses the computed probabilities to select a new eighth token, and then generates 2 more tokens to obtain 10 tokens in all, the generation ratio is 1.1 (11 model invocations for 10 output tokens).

We note that for the BigCodeBench results, a data error caused the denominator of the generation ratio to reflect the number of tokens in the output after converting from tokens, to text, back to tokens, rather than reflecting the original number of generated tokens. 
This causes the generation ratios to vary slightly from their true value (usually by 0.001-0.005); to be overly conservative, we truncate the ratio to two significant digits after the decimal.
Ratios for other experiments are not affected by this.

\section{Extended Example of Probability Amplification}

\label{appendix:extended-probability-amplification}

\begin{figure}
    \centering
    \begin{subfigure}{0.5\textwidth}
        \includegraphics[width=0.9\linewidth,page=7]{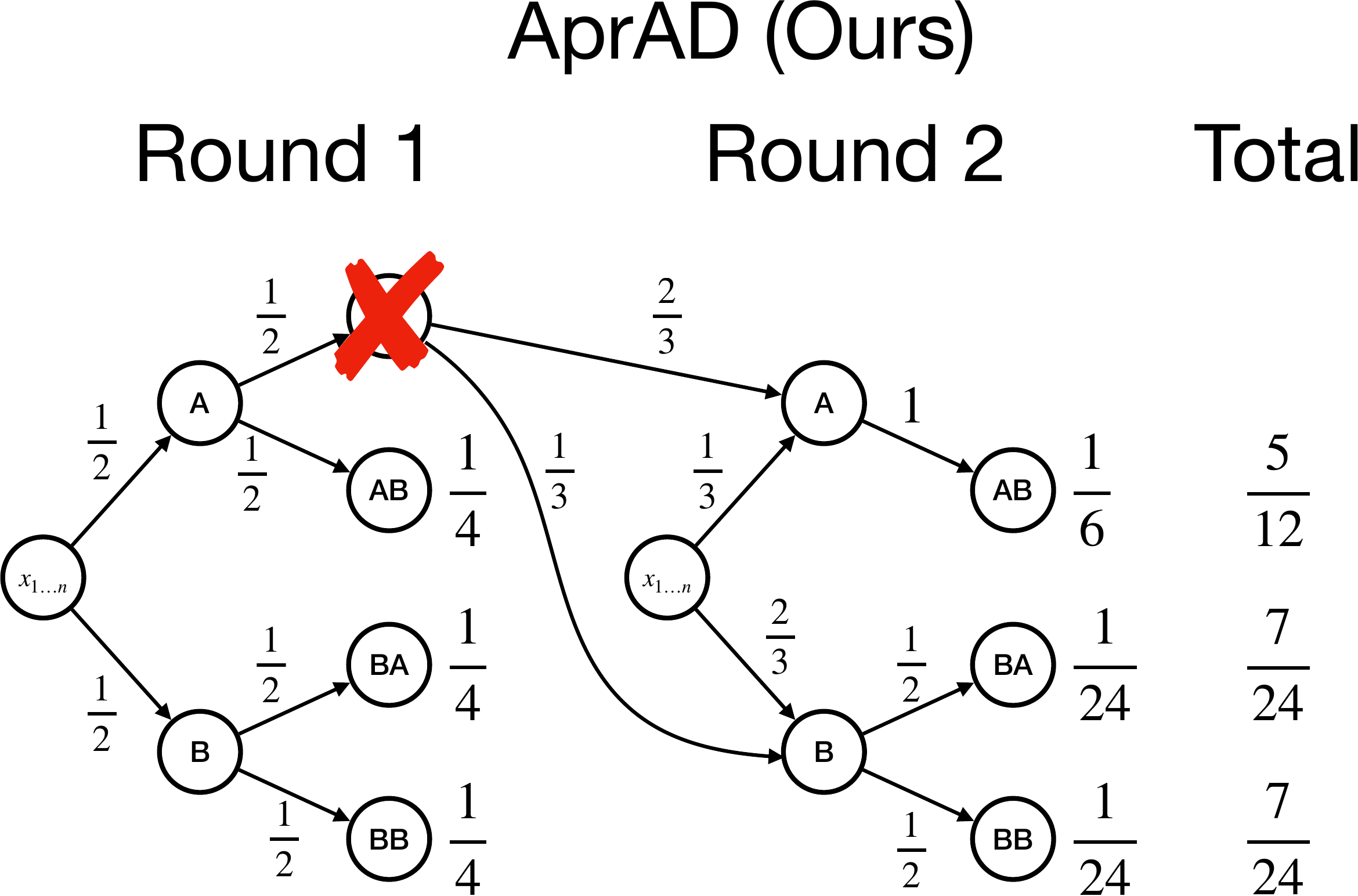}    \caption{AprAD Probability Distribution (from Figure \ref{fig:run-example-aprad}).}
    \end{subfigure}%
    \begin{subfigure}{0.5\textwidth}
            \includegraphics[width=0.9\linewidth,page=9]{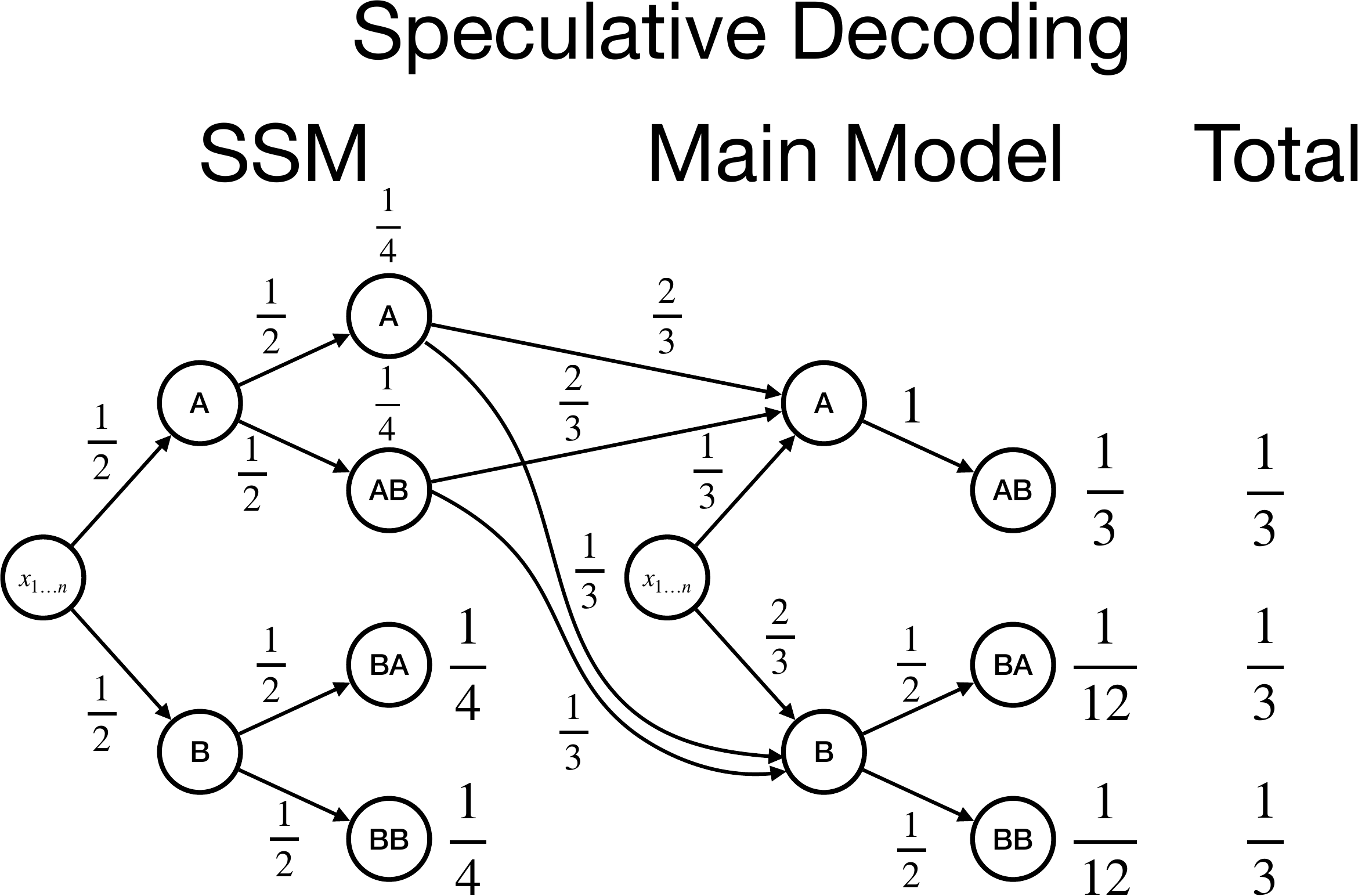}    \caption{Speculative Decoding Probability Distribution.}
    \end{subfigure}
    \caption{Comparison of probability distribution between AprAD and Speculative Decoding}
    \label{fig:comparison-probability-distribution}
\end{figure}

We further illustrate the cause of probability amplification, despite using the speculative decoding algorithm, through an extended example.

Consider the running example in Figure \ref{fig:comparison-probability-distribution}.
In the first round, there is a $\frac{1}{4}$ probability each of selecting sequence AA, AB, BA, or BB.

In the case where AA is selected, AprAD has identical behavior to speculative decoding where $\llm$ is the speculative model and $\hat{\llm}^{\{\text{AA}\}}$ is the main model; \textsc{SpecSample} is invoked in both cases.
We note that in cases where $|\badSet| > 1$, after the first round, AprAD has still only observed one element $x$ of $\badSet$ and thus must invoke \textsc{SpecSample} with $\hat{\llm}^{\{x\}}$ rather than with $\llmExcBadSet$.

In the case where AB is selected, the behavior is completely different. 
With speculative decoding, it is still the case that $\llmExcBadSet(A) < \llm(A)$, leading to a potential backtrack.
With AprAD however, since a non-violating sequence is found, the process terminates---the process has no mechanism to discover $\llmExcBadSet$ other than through encounters with elements of $\badSet$.

With both BA and BB, the speculative decoding process accepts the SSM's output because $\llmExcBadSet(B) \geq \llm(B)$, $\llmExcBadSet(A|B) \geq \llm(A|B)$, and , $\llmExcBadSet(B|B) \geq \llm(B|B)$.
AprAD also accepts these sequences immediately, as both sequences are non-violating.

\textbf{Summary.} The probability amplification of AprAD occurs due to the cases where \textsc{SpecSample} is \textit{not} invoked.

\section{Bounding the Probability Amplification Factor}

\label{appendix:probability-amplification-bound}

While the following sketch is not a rigorous proof, we argue that the probability amplification due to AprAD is at most $2$ per iteration of the algorithm.

For sequences $x = x_1, \ldots, x_n$ and $y = y_1, \ldots, y_m$, let $SRS(y|x, S, P)$ (SpeculativeReSample) be the probability of eventually selecting sequence $y$ with speculative decoding, conditional on having drawn $x$ from speculative model $S$.

\textbf{Speculative Decoding Identity.} The original concept of speculative decoding relies on the identity $P(y) = \sum_{x \in \Sigma^*} S(x) SRS(y|x, S, P)$---the probability of selecting $y$ with the main model $P$ should be equal to the probability of selecting $y$ using the speculative decoding process.

\textbf{SRS Subset Inequality.} If $\badSeen \subseteq \badSet$, then $SRS(y|x, S, \llmExcBadSeen) \leq SRS(y|x, S, \llmExcBadSet)$ for $y \notin \badSet$. Intuitively, $\badSet$ excludes more invalid sequences compared to $\llmExcBadSeen$, so the probability mass of these sequences during the re-sampling process should be distributed among all non-error sequences, including $y$.

\textbf{Generation Subset Inequality.} For similar reasons, $\llmExcBadSeen(x) \leq \llmExcBadSet(x)$ for $\badSeen \subseteq \badSet, x \notin \badSet$, as excluding additional errors $\badSet \setminus \badSeen$ and re-normalizing the probability means that all other sequences become more likely.

The probability of sequence $y \notin \badSet$ being generated after the first iteration of AprAD is equal to the probability that $y$ is generated directly, plus the probability that $y$ is re-sampled after some invalid sequence is generated:

$$AprAD(y|P,\badSet) = P(y) + \sum_{x \in \badSet} P(x) SRS(y|x, P, \hat{P}^{\{x\}})$$

We apply both of the subset inequalities described above:

$$\leq \llmExcBadSet(y) + \sum_{x \in \badSet} P(x) SRS(y|x, P, \llmExcBadSet)$$

As $P(x)$ and $SRS(y|x, P, \llmExcBadSet)$ are always nonnegative, we can add additional elements to the sum---expanding it to include all sequences, rather than just error sequences---while maintaining the inequality:

$$\leq \llmExcBadSet(y) + \sum_{x \in \Sigma^*} P(x) SRS(y|x, P, \llmExcBadSet)$$

And finally use the speculative decoding identity:

$$\leq 2 \llmExcBadSet(y)$$

Note that this amplification is per-iteration, where an iteration is defined as encountering an error sequence, potentially backtracking, and resampling. In practice, the per-iteration amplification is likely much less, as several of the inequalities involved are very loose. 
However, there may be a cumulative effect as more iterations are required in dense error sets.

In contrast, the probability amplification after encountering even a single error sequence with constrained decoding is unbounded, as the best non-error token may have an arbitrarily low probability.

\section{Additional BigCodeBench Results}

\begin{table*}[ht]
    \centering
    \small
    \begin{tabular}{ccccccc}\toprule
    Size & Method & Pass@1 & Pass@5 & !NameErr@1 & !NameErr@5 & Gen. Ratio \\ \midrule
\multirow{4}{*}{15b}& Unconstrained & 0.31 $\pm$ 0.01 & 0.58 $\pm$ 0.01  & 0.95 $\pm$ 0.00 & 0.99 $\pm$ 0.00& \textbf{1.00 $\pm$ 0.00} \\
& AprAD (Ours) & \textbf{0.32 $\pm$ 0.01} & \textbf{0.59 $\pm$ 0.01 } & \textbf{0.98 $\pm$ 0.00} & \textbf{1.00 $\pm$ 0.00} & 1.02 $\pm$ 0.00 \\
& ASAp & \textbf{0.32 $\pm$ 0.01} & \textbf{0.59 $\pm$ 0.01}  &\textbf{ 0.98 $\pm$ 0.00} & \textbf{1.00 $\pm$ 0.00} & 1.11 $\pm$ 0.02 \\
& Constrained & 0.31 $\pm$ 0.01 & 0.58 $\pm$ 0.01  & 0.97 $\pm$ 0.00 & \textbf{1.00 $\pm$ 0.00 }&\textbf{ 1.00 $\pm$ 0.00} \\\midrule
\multirow{4}{*}{7b}& Unconstrained & 0.20 $\pm$ 0.01 & 0.47 $\pm$ 0.01  & 0.93 $\pm$ 0.00 & \textbf{0.99 $\pm$ 0.00} & \textbf{1.00 $\pm$ 0.00} \\
& AprAD (Ours) & \textbf{0.21 $\pm$ 0.01} & 0.47 $\pm$ 0.01  & \textbf{0.97 $\pm$ 0.00} & \textbf{0.99 $\pm$ 0.00} & 1.02 $\pm$ 0.00 \\
& ASAp & \textbf{0.21 $\pm$ 0.01} & \textbf{0.48 $\pm$ 0.01}  & \textbf{0.97 $\pm$ 0.00 }& \textbf{0.99 $\pm$ 0.00} & 1.13 $\pm$ 0.02 \\
& Constrained & 0.20 $\pm$ 0.01 & 0.47 $\pm$ 0.01  & 0.95 $\pm$ 0.00 & \textbf{0.99 $\pm$ 0.00} & 1.00 $\pm$ 0.00 \\\bottomrule
    \end{tabular}
    \caption{Results for each method on entirety of BigCodeBench. Note that these results are identical to those in Table \ref{tab:bigcodebench_only_different}, except that they are consistently offset and scaled to include values for tasks in which all tasks return the same result.}
    \label{tab:bigcodebench_all_results}
\end{table*}

Table \ref{tab:bigcodebench_all_results} includes results for the entirety of BigCodeBench; not just the tasks for which the methods diverged in their output.

\section{Generalization of Error-Free Decoding}

Constrained generation, ASAp, and AprAD may all be generalized by their backtracking behavior after an error is discovered.
Algorithm \ref{algo:error-free-generalization} shows this generalization.

\label{section:error-free-generalization}

\begin{algorithm}
\begin{algorithmic}
\Procedure{ErrFreeDec}{$\llm, \badSet, x_{1\ldots n}, \textsc{Strategy}$}
    \State $\llmExcBadSeen \leftarrow \llm$
    \State $m \leftarrow n$  \Comment{Current token index}
    \While{Stopping condition not met}
        \State Sample one token $x_{m+1} \sim \llmExcBadSeen(\cdot | x_{1\ldots m})$
        \State Increment $m$
        \If{$x_{1\ldots m}  \in \badSet$}
         \LComment{Algorithm \ref{algo:asap}}
            \State $\llmExcBadSeenAndX \leftarrow \Call{AddBadSample}{\llmExcBadSeen, x_{1\ldots m}}$
            \LComment{$m$ may decrease}
            \State $x_{1\ldots m} \leftarrow \Call{Strategy}{\llmExcBadSeen, \llmExcBadSeenAndX, x_{1\ldots m}}$ 
            \State $\llmExcBadSeen \leftarrow \llmExcBadSeenAndX$
        \EndIf
    \EndWhile
    \State \Return $x_{1\ldots m}$
\EndProcedure
\Procedure{AprADStrategy}{$\llmExcBadSeen, \llmExcBadSeenAndX, x_{1\ldots m}$}
\LComment{Algorithm \ref{algo:specsample}}
  \State \Return \Call{SpecSample}{$\llmExcBadSeen, \llmExcBadSeenAndX, 0, x_{1\ldots m}$} 
\EndProcedure
\Procedure{ASApStrategy}{$\llmExcBadSeen, \llmExcBadSeenAndX, x_{1\ldots m}$}
  \State \Return [] \Comment{Backtrack to beginning}
\EndProcedure
\Procedure{ConstrainedDecodingStrategy}{$\llmExcBadSeen, \llmExcBadSeenAndX, x_{1\ldots m}$}
\LComment{Delete error token but don't backtrack further}
  \State \Return $x_{1\ldots m-1}$ 
\EndProcedure
\end{algorithmic}
\caption{Multiple error-free decoding methods differ only in their backtracking selection.}
\label{algo:error-free-generalization}
\end{algorithm}

\section{Method Comparisons}

\begin{table*}[t]
    \centering
    \scriptsize
    \begin{threeparttable}
    \begin{tabular}{c|c|c|c|c}
       Method  & Overhead $\downarrow$ & Dist. Conformance $\uparrow$ & Constraint Class & Posterior Estimate \\\hline
       AprAD (Ours)  & Medium & Medium & Black Box\tnote{a} & Not Required \\
       ASAp \citep{park2024grammaraligneddecoding} & High & High & Black Box\tnote{a} & Not Required \\
       Constrained (Multiple) & Low & Low & Black Box\tnote{a} & Not Required\\
       FUDGE \citep{FUDGE_Yang_2021} & Low\tnote{b} & High\tnote{c} & Prefix-Dependent\tnote{d} & Learned Discriminator \\
       SMC Steering \citep{lew2023sequentialmontecarlosteering} & High & High & Black Box\tnote{a} & Sample Many Rollouts \\
       Ctrl-G \citep{zhang2024adaptablelogicalcontrollarge} & Low\tnote{b} & High\tnote{c} & Represent as DFA & Exact in distilled HMM
    \end{tabular}
    \begin{tablenotes}
        \item [a] Oracle classifies whether a given output violates constraint.
        \item [b] Requires additional one-time training step per task.
        \item [c] Conditional on learned or distilled model perfectly capturing LLM behavior.
        \item [d] Requires that it is possible to determine probability of constraint violation from incomplete prefix.
    \end{tablenotes}
    \end{threeparttable}
    \caption{Overview of several methods for controllable generation with a LLM, with subjective estimate of inference overhead and conformance to the LLM's original output distribution, and a brief description of constraint expressivity and method of posterior estimation.}
    \label{tab:method-comparison}
\end{table*}

An overview comparison of several methods is presented in Table \ref{tab:method-comparison}.

\section{Compute Resources}
\label{appendix:resources}

Computation time was not precisely tracked during preliminary or final experiments; several experiments ran in parallel, resulting in a high variation of inference speed. 
Inference time was dominated by the BigCodeBench trials (1140 tasks, 4 methods, 2 models, 5 trials per model = 45600 generations. 
Order of magnitude of $10^3$ output tokens per trial, resulting in about $10^7-10^8$ output tokens).
This took about a day or two on an AWS p4d.24xlarge instance.
The lipogram task was comparatively much less resource intensive, with 25 prompts, 4 models, and 200 output tokens per prompt, with a higher generation ratio than BigCodeBench, resulting in about $10^4-10^5$ model invocations.
The simulated testbench experiments run in seconds to minutes on a consumer laptop.

\section{Implementation Details}
\label{section:implementation_details}

We provide notes related to our implementation of the methods discussed in this paper.
In particular, floating point error accumulation was a major source of issues due to invariants breaking late in the generation process; we note where recalculations may be required to avoid this.

\subsection{ASAp}

An implementation of ASAp is provided in Algorithm \ref{algo:asap}.

\begin{algorithm}
\begin{algorithmic}
\Procedure{ASAp}{$\llm, \badSet, x_{1\ldots n}$} \Comment{$x_{1\ldots n}$ is prompt}
    \State $\llmExcBadSeen \leftarrow \llm$
    \While{Limit not reached}
        \State Sample sequence $x_{n+1\ldots m} \sim \llmExcBadSeen(\cdot | x_{1\ldots n})$ until error or stopping condition
        \If{$x_{1\ldots m} \notin \badSet$} \textbf{break} \EndIf
        \LComment{Remove $x_{1\ldots m}$ as a possible sequence}
        \State $\llmExcBadSeen \leftarrow \text{AddBadSample}{\llmExcBadSeen, x_{1\ldots m}}$ 
    \EndWhile
    \State \Return $X$
\EndProcedure
\end{algorithmic}
\caption{ASAp \citep{park2024grammaraligneddecoding}}
\label{algo:asap}
\end{algorithm}

\subsection{Trie-Structured Probability Cache, \textsc{AddBadSample}, and Cached Probabilities}

After each token probability distribution is generated from the language model, we add it to a trie structure to represent $\llmExcBadSeen$ efficiently.

The node representing prefix $x_{1\ldots m}$ contains the following:
\begin{itemize}
    \item A single token $x_m$, and a pointer to a parent node representing $x_{1\ldots {m-1}}$
    \item The original probabilities generated by the LLM $\llm(\cdot | x_{1\ldots m})$.
    \item The modified conditional probabilities $\llmExcBadSeen(\cdot | x_{1\ldots m})$.
    \begin{itemize}
        \item Due to floating point implementation issues, and for efficiency purposes, we store these modified probabilities un-normalized; i.e. we store a table $\llmExcBadSeen{}^*(\cdot | x_{1 \ldots m})$ where $\sum_{x_{m+1} \in \Sigma} \llmExcBadSeen{}^*(x_{m+1} | x_{1 \ldots m}) \leq 1$. 
        \item We track this sum in a variable, $f$, and divide the un-normalized probabilities by $f$ as necessary to obtain normalized probabilities when queried.
        When $f$ is small, and likely to suffer from accumulated floating point errors, we periodically recalculate it by summing the $\llmExcBadSeen{}^*$ table.
        \item Additionally, when an entry of $\llmExcBadSeen{}^*$ is sufficiently small, or becomes negative, we assume that its value is zero, but has excessively suffered from accumulated floating point errors; we therefore set it to zero.
    \end{itemize}
\end{itemize}

This structure allows for an efficient implementation of \textsc{AddBadSample}, as given in Algorithm \ref{algo:implementation_add_bad_sample}. 
The same trie structure is also used to track the adjusted probabilities for the comparison methods.

\begin{algorithm}
\begin{algorithmic}
\Procedure{AddBadSample}{Node $n$} \LComment{$n$ represents a violating sample}
    \State $t \leftarrow n.token$
    \State $c \leftarrow n.parent$ \Comment{Current node}
    \State $r \leftarrow 1$
    \While{$c$ is not null} 
        \LComment{Calculate $r$ as probability of violating sample in $c$, in original distribution}
        \State $r \leftarrow r \times c.\llm(t)$
        \State $c.\llmExcBadSeen{}^*(t) \leftarrow c.\llmExcBadSeen{}^*(t) - r$
        \State $c.f \leftarrow c.f - r$ \Comment{Recalculate $c.f$ as sum of $c.\llmExcBadSeen{}^*$ if needed due to FP errors}
        \State $t \leftarrow c.token$
        \State $c \leftarrow c.parent$
    \EndWhile
\EndProcedure
\end{algorithmic}
\caption{An implementation-oriented description of \textsc{AddBadSample}}
\label{algo:implementation_add_bad_sample}
\end{algorithm}

\subsection{Backtracking Strategies}

\label{section:backtracking-strategies}

Our implementation uses essentially the same structure as detailed in Appendix \ref{section:error-free-generalization}, where it is parameterized by a backtracking strategy.
However, it would be complex to maintain full probability tries representing both $\llmExcBadSeen$ and $\llmExcBadSeenAndX$ every time a violating sample is found.
We observe that it is unnecessary to do so; rather, we only need both probabilities along the ``path'' of the violating sample; i.e. $\llmExcBadSeen(x_1), \llmExcBadSeen(x_2 | x_1), \ldots \llmExcBadSeen(x_m | x_{1\ldots m-1})$.
Therefore, we invoke $\textsc{CollectProbabilities}$ (Algorithm \ref{algo:collect_probabilities}) before calling $\textsc{AddBadSample}$.

\begin{algorithm}
\begin{algorithmic}
\Procedure{CollectProbabilities}{Node $n$} 
    \State $t \leftarrow n.token$
    \State $c \leftarrow n.parent$ \Comment{Current node}
    \State $o \leftarrow []$
    \While{$c$ is not null} 
        \LComment{Calculate $r$ as probability of violating sample in $c$, in original distribution}
        \State $\textsc{Append}(o, c.f \times c.\llmExcBadSeen{}^*(t))$
        \State $t \leftarrow c.token$
        \State $c \leftarrow c.parent$
    \EndWhile
    \State \Return $\textsc{Reverse}(o)$
\EndProcedure
\end{algorithmic}
\caption{Collect all probabilities along a path}
\label{algo:collect_probabilities}
\end{algorithm}

The $\textsc{SpecSample}$ algorithm is used nearly unmodified, except that $P$ is given by the adjusted conditional probabilities in the trie, and $S$ is given by the cached list collected by \textsc{CollectProbabilities}.

We make one modification to constrained decoding, necessary due to our use of top-p sampling such that there may not always be an available token that satisfies the constraint in the next step.
If the current leaf node of the probability trie has no non-zero entries in $\llmExcBadSeen{}^*$, then our implementation of constrained decoding will backtrack until there is at least one available continuation.

\section{Additional Lipogram Examples}
\label{section:more_lipogram_examples}

The following pages contain generations from 10 random prompts of the lipogram task. 
The examples are not cherry-picked.
If the banned letter appears---this occurs in all unconstrained examples, and no other method---it is \textbf{bolded}.
Non-ASCII characters are underlined and highlighted in \rus{red}.
We include </s> where the language model chooses to stop generation; otherwise, it was cut off after 200 tokens or 2000 model invocations, whichever occurred first.
Generations from all prompts are available in the supplemental material.

        \genBox{\textbf{Provide instructions to tie a tie without using the letter "I".}}{\textbf{AprAD (Ours)}}{To create an elegant neckwear look, follow these steps to adorn your collar:

1. Gather your tools: Choose a well- Press your collar down and adjust the top button.

2. Select the perfect knot: There are many knots to choose from; select a preferred one.

3. Dress your end: Take the end nearest to you and form a loop, called the "small" or "narrow" end.

4. Create an overlap: Cross the longer end, called the "large" end, over the loop.

5. Form a second loop: Take the small end and make another loop.

6. Create a "V" shape: Place the large end over the two loops. Adjust for symmetry.

7. Repeat: Cross the small end over the large end, then the large end over the small end to create a "V" shape once more.

8. Tuck and adjust: T}

        \genBox{\textbf{Provide instructions to tie a tie without using the letter "I".}}{Unconstrained}{To create an elegant neckt\textbf{i}e knot, follow these steps:

1. Lay the t\textbf{i}e flat on the table w\textbf{i}th the w\textbf{i}de end (the end you'll be ty\textbf{i}ng f\textbf{i}rst) fac\textbf{i}ng you. The th\textbf{i}n end (the s\textbf{i}de you'll be attach\textbf{i}ng to thesh\textbf{i}rt) should be hang\textbf{i}ng on the left.

2. Cross the w\textbf{i}de end over the th\textbf{i}n end and br\textbf{i}ng \textbf{i}t beh\textbf{i}nd the looped th\textbf{i}n end.

3. Now, pass the w\textbf{i}de end over the looped th\textbf{i}n end from left to r\textbf{i}ght, go\textbf{i}ng around the back of the neckt\textbf{i}e.

4. Keep\textbf{i}ng the w\textbf{i}de end beh\textbf{i}nd the looped th\textbf{i}n end, cross \textbf{i}t over to the left s\textbf{i}de and br\textbf{i}ng \textbf{i}t \textbf{i}n front of the looped end.

5. Now, the w\textbf{i}de end should be on the left and the looped end on the r\textbf{i}ght. Cross the w\textbf{i}de end over the looped end aga\textbf{i}n to the r\textbf{i}ght s\textbf{i}de.

6. Pass the w\textbf{i}de end under}

        \genBox{\textbf{Provide instructions to tie a tie without using the letter "I".}}{ASAp}{To create}

        \genBox{\textbf{Provide instructions to tie a tie without using the letter "I".}}{Constrained}{To create an elegant necktube knot for your formal att attar, follow these steps:

1. Lay the necktube flat on the table, end towards you.

2. Fold one end about four to four and a half \rus{и}\rus{н}ches (10-12 cm) over the other end.

3. Create a loop, called the "small loop," by fold the part of the necktube above the folded end, about two to three \rus{и}\rus{н}ches (5-7.5 cm) down.

4. Create a second loop, called the "large loop," by fold the part of the necktube around your neck, and rest the large loop over the small loop.

5. Tuck the end of the necktube under the large loop, then the small loop.

6. Pass the end of the necktube under the small loop and through the hole created by the knot.

7. Tug on}
        
\newpage

        \genBox{\textbf{Provide instructions to tie a tie without using the letter "O".}}{\textbf{AprAD (Ours)}}{1. Begin with the tie unwinded and the widest end (the "bracelet end") tucked under the neckband.

2. Bring the wider end (the "tie end") behind the neckband, then tuck it under the bracelet end and pin it in place with a thin fabric "shelf" called the "keeper."

3. Create a "Wing Tip" by lifting the neckband and slipping the tie end under it, then pull it up evenly and adjust the width until it's level with the bracelet end.

4. With the tie end still held in place with the neckband, create a "Shelf" by lifting the bracelet end and passing the tie end beneath it, then adjust the tie width again by pulling the tie end evenly.

5. Create a "Three-inch rule" by sliding the bracelet end up three inches (7.6 cm}

        \genBox{\textbf{Provide instructions to tie a tie without using the letter "O".}}{Unconstrained}{1. Begin by laying \textbf{o}ut y\textbf{o}ur tie \textbf{o}n a flat surface with the wide end, kn\textbf{o}wn as the "tail," facing y\textbf{o}u. The narr\textbf{o}wer end is called the "Tip."

2. Cr\textbf{o}ss the Tip \textbf{o}ver the Tail fr\textbf{o}m left t\textbf{o} right.

3. Bring the Tail ar\textbf{o}und the back \textbf{o}f the neck and cr\textbf{o}ss it \textbf{o}ver the Tip.

4. N\textbf{o}w, take the Tip and thread it under the l\textbf{o}\textbf{o}p created by the Tail.

5. Pull the Tip up thr\textbf{o}ugh the l\textbf{o}\textbf{o}p.

6. Adjust the p\textbf{o}siti\textbf{o}n \textbf{o}f the Tip and Tail t\textbf{o} achieve a symmetrical kn\textbf{o}t.

7. Repeat the pr\textbf{o}cess \textbf{o}n the \textbf{o}ther side, cr\textbf{o}ssing the Tip \textbf{o}ver the Tail, bringing the Tail ar\textbf{o}und the back, threading the Tip under the l\textbf{o}\textbf{o}p, and pulling it thr\textbf{o}ugh.

8. With b\textbf{o}th sides adjusted, adjust the kn\textbf{o}t \textbf{o}ne final time t\textbf{o} ensure a neat and symmetrical l\textbf{o}\textbf{o}k.</s>}

        \genBox{\textbf{Provide instructions to tie a tie without using the letter "O".}}{ASAp}{1}

        \genBox{\textbf{Provide instructions to tie a tie without using the letter "O".}}{Constrained}{1. Begin by laying the necktie flat in a figure eight shape. The wide end (called the "butt end") will be anchered at the left side, and the thin end (called the "tip end") will be adjusted at the right side.

2. Bring the right side (the side with the tip end) up and behind the left side, creating a "U" shape, then pull it straight up and tuck the tip end under the wide part at the nape (the back part) near the center.

3. Next, bring the left side up and pass it (the wide end) behind the neck, then pull it straight up and rest it against the right side.

4. With the left side still in hand, take the right side (tip end) and bring it up and pass it behind the left side.

5. Tuck the tip end behind and under the wide part (where the}
        
\newpage

        \genBox{\textbf{Write a story without using the letter "E".}}{\textbf{AprAD (Ours)}}{In a quaint, old-world town, a mystical story unfolds. A solitary knight, Sir Gavr, hails from a distant land. His mission was vital, as his king had sought his aid. A monstrous dragon, with a long, sinuous form, had laid claim to a bountiful land.

Sir Gavr, galloping through twisting hills and past lilac valms, his stallion, Swiftpaw, carrying him swiftly. Through woodlands dark and mossy, a path narrow and winding. His armor glinting gold, a radiant sight, as knight and mount, in rhythmic flight, cut through air, crisp and light.

A labyrinth of hills and hollows, Sir Gavr and Swiftpaw, onward bound. A looming mountain, a daunting sight, as dragon's lair, a}

        \genBox{\textbf{Write a story without using the letter "E".}}{Unconstrained}{In a quaint, old-world villag\textbf{e} liv\textbf{e}d a wis\textbf{e} artisan nam\textbf{e}d Gobo. His humbl\textbf{e} abod\textbf{e} was n\textbf{e}stl\textbf{e}d 'twixt a grov\textbf{e} of tow\textbf{e}ring, v\textbf{e}rdant tr\textbf{e}\textbf{e}s. His days w\textbf{e}r\textbf{e} pass\textbf{e}d in crafting intricat\textbf{e}, \textbf{e}xquisit\textbf{e} pott\textbf{e}ry pi\textbf{e}c\textbf{e}s, \textbf{e}ach mor\textbf{e} b\textbf{e}autiful than th\textbf{e} last.

Gobo sourc\textbf{e}d clay from a n\textbf{e}arby str\textbf{e}am, wh\textbf{e}r\textbf{e} its cool, crystallin\textbf{e} wat\textbf{e}rs danc\textbf{e}d ov\textbf{e}r smooth, mossy p\textbf{e}bbl\textbf{e}s. Th\textbf{e} sun, a gold\textbf{e}n orb, bath\textbf{e}d th\textbf{e} clay in warm hu\textbf{e}s b\textbf{e}for\textbf{e} Gobo shap\textbf{e}d it with skill\textbf{e}d, nimbl\textbf{e} fing\textbf{e}rs. A wh\textbf{e}\textbf{e}l, spun by muscular arms, transform\textbf{e}d th\textbf{e} mall\textbf{e}abl\textbf{e} substanc\textbf{e} into symm\textbf{e}trical rounds.

Onc\textbf{e} shap\textbf{e}d, pi\textbf{e}c\textbf{e}s w\textbf{e}r\textbf{e} s\textbf{e}t to dry und\textbf{e}r th\textbf{e} cl\textbf{e}ar, azur\textbf{e} sky. With pati\textbf{e}nc\textbf{e}, th\textbf{e}y transform\textbf{e}d into d\textbf{e}licat\textbf{e}, hard\textbf{e}n\textbf{e}d cr\textbf{e}ations. With d\textbf{e}ft strok\textbf{e}s, Gobo adorn\textbf{e}d pots and bowls with intricat\textbf{e},}

        \genBox{\textbf{Write a story without using the letter "E".}}{ASAp}{I.}

        \genBox{\textbf{Write a story without using the letter "E".}}{Constrained}{In a quaint, old-world town, N, a kindly cobbl\rus{е}\rus{р}, dwlt with his donk mulk. His daily task was to mak sho'd that N's villagrs' shojs staid sound. With nimbl fingrts and skillful hand, N smotd, nwd, and polisht prgrssvly.

Bys N's humbl abod, a gldng fountain splshd light thru wndw shds. It was a gif frm hm t Ols N., N's fav or old fnd, who hd lvd n ths tow n yrs ag.

Today, though, N fnd dffrlt to wk. Hs thmb, hnd, awoy aggrv, paining hm wth sharp, stng p}
        
\newpage

        \genBox{\textbf{Describe elephants without using the letter "O".}}{\textbf{AprAD (Ours)}}{Elephants are large, gray, thick-skinned pachyderms, characteristically having extended, curved tusks, and a fan-shaped, large ear structure. They are the largest living terrestrial animals, with a distinct, massive build. These gentle giants have a wrinkled visage, and large, dangling, gray, prehensile ears, which are effective at dissipating heat. The trunk, a lengthy, muscular, and versatile appendage, is an integral feature, serving multiples uses, including breathing, grasping, and trumpeting. Their legs are pillar-like, ending in large, flat feet, which help them maneuver in diverse terrains. Their size and strength make them apex terrestrial animals, and they are crucial elements in their respective habitats.</s>}

        \genBox{\textbf{Describe elephants without using the letter "O".}}{Unconstrained}{Elephants are large, gray, thick-skinned pachyderms, kn\textbf{o}wn f\textbf{o}r their l\textbf{o}ng, curved tusks. They are the largest land animals, reaching up t\textbf{o} 13 feet in length and 10 feet in height at the sh\textbf{o}ulder. With str\textbf{o}ng, sturdy legs, they supp\textbf{o}rt up t\textbf{o} 13,000 p\textbf{o}unds. Their el\textbf{o}ngated trunk, a flexible extensi\textbf{o}n \textbf{o}f their upper lip, is used f\textbf{o}r tasks including grasping f\textbf{o}\textbf{o}d and water, trumpeting, and bathing. Their ears, which functi\textbf{o}n in b\textbf{o}th temperature regulati\textbf{o}n and c\textbf{o}mmunicati\textbf{o}n, can reach up t\textbf{o} 5 feet in width. These gentle giants live in herds, displaying c\textbf{o}mplex s\textbf{o}cial behavi\textbf{o}rs and intricate hierarchies.</s>}

        \genBox{\textbf{Describe elephants without using the letter "O".}}{ASAp}{Large, gray, thick-sk}

        \genBox{\textbf{Describe elephants without using the letter "O".}}{Constrained}{Elephants are large, gray, thick-skinned pachyderms, inhabiting African and Asian savannas, rain-drenched jungles, and diverse habitats in between. Kneeling, they rest their massive, curved tusks, used primarily in feeding and defense, against the jungle's underbrush. Their elastic, dexterity-laden trunks, extended appendages, twist and twirl, grabbing leaves, branches, and fruits. Their bulbus, enlarged, sensitive tips, caress and manipulate their quenching water and tender infants. Males and females, banded, marked, and tusked, live in matriarchal, familial, and herd-like structures, their distinct, rumbling rumbles, a testament, that despite their gargantuan frames, these gentle giants are an integral part, in the intr}
        
\newpage

        \genBox{\textbf{Critique the Mona Lisa without using the letter "E".}}{\textbf{AprAD (Ours)}}{In this artwork, a captivating woman, known as La Joconda or La Hora, holds a puzzling grin. A portrait, it is, of this cryptic woman. With a backdrop of lush, rolling hills, this painting is a symphony of color and form.

This woman, clad in a rich, royal gown, sits in a chair in an unusual position, twisting slightly, inviting an intimacy that is both alluring and intriguing. A slight smirk, or is it a sly grin? This ambiguity is what draws us in, as if to ask, "Am I amusing or mocking?"

Boldly, shadows and light play in this work, as if dancing a waltz, with chiaroscuro as its music. This contrast, striking and dramatic, is what adds an almost magical quality to this artwork.

A mosaic of colors, this painting is}

        \genBox{\textbf{Critique the Mona Lisa without using the letter "E".}}{Unconstrained}{Th\textbf{e} Mona Lisa, a singular 16th-c\textbf{e}ntury painting, is L\textbf{e}oni da Vinci's \textbf{e}pitom\textbf{e}. This mast\textbf{e}rpi\textbf{e}c\textbf{e}, p\textbf{e}nn\textbf{e}d by th\textbf{e} virtuoso artist, \textbf{e}vok\textbf{e}s a s\textbf{e}ns\textbf{e} of \textbf{e}nigma and captivation. Th\textbf{e} woman portray\textbf{e}d, with h\textbf{e}r \textbf{e}nigmatic smil\textbf{e}, gaz\textbf{e}s dir\textbf{e}ctly at th\textbf{e} obs\textbf{e}rv\textbf{e}r, chall\textbf{e}nging on\textbf{e}'s int\textbf{e}rpr\textbf{e}tation.

Th\textbf{e} painting's composition mast\textbf{e}rfully balanc\textbf{e}s th\textbf{e} human form against a lush background, showcasing L\textbf{e}onardo's virtuosic skill. Th\textbf{e} woman, dr\textbf{e}ss\textbf{e}d in a subtl\textbf{e}, intricat\textbf{e} rob\textbf{e}, is contrast\textbf{e}d by th\textbf{e} stark, plain backdrop. H\textbf{e}r \textbf{e}xpr\textbf{e}ssion, subtly shifting b\textbf{e}tw\textbf{e}\textbf{e}n coy and cont\textbf{e}mplativ\textbf{e}, invit\textbf{e}s \textbf{e}ndl\textbf{e}ss analysis.

Th\textbf{e} intricat\textbf{e} play of light and shadow on h\textbf{e}r fac\textbf{e} adds d\textbf{e}pth and dim\textbf{e}nsion, as if th\textbf{e} painting its\textbf{e}lf holds a s\textbf{e}cr\textbf{e}t. Th\textbf{e} artist's ad\textbf{e}pt us\textbf{e} of sfumato t\textbf{e}chniqu\textbf{e}, wh\textbf{e}r\textbf{e} colors and ton\textbf{e}s bl\textbf{e}nd}

        \genBox{\textbf{Critique the Mona Lisa without using the letter "E".}}{ASAp}{In this}

        \genBox{\textbf{Critique the Mona Lisa without using the letter "E".}}{Constrained}{I'll do my utmost to critic this Mona Lisa without utilizing "\rus{é}" or any of its variants, including "\rus{é}" as in "\rus{é}tudiant" or "\rus{è}" as in "m\rus{è}asuring." I'll focus on "critic" as in "judging with an analytical and authoritarian approach."

This painting, La Gioconda, by da Vinci, is an intriguing, cryptic work of art. Its woman's countanza, or count\rus{é}nanc, or visagio, or visagio da vita, or simply, its facia, is a captivating, almost hypnotizing, sight.

With a sfumato background that dissipatas, or dissipas, into an indistinct, unbound horizon, our ojo is drawn to its suj\rus{е}\rus{т}\rus{о} principal. Its sfumato, or "smoky," background is in striking contrast}
        
\newpage

        \genBox{\textbf{Critique the Mona Lisa without using the letter "A".}}{\textbf{AprAD (Ours)}}{Leonymus L\rus{é}onifor de Vinci's iconic piece, the 'Lizone Jocundity' or 'Smiling Lioness,' epitomizes the refined intrigue of the 16th century. This work, completed in the mid-1500s, is renowned for its subject's elusive expression, veiled in mystery.

The 'Lizone' is executed with the finest of techniques. Leonifor's virtuosity is evident in the meticulous, lifelike rendering of the figure's physiognomy, dressed in the subdued splendor of the epoch. The sitter, shrouded in the mists of time, seems to defy definition.

The composition is striking; the figure is posed, enshrouded in the folds of her veil, the folds of her robe - the folds of her being.}

        \genBox{\textbf{Critique the Mona Lisa without using the letter "A".}}{Unconstrained}{The Mon\textbf{a} Lis\textbf{a}, \textbf{a} iconic m\textbf{a}sterpiece by Leon\textbf{a}rdo d\textbf{a} Vinci, is \textbf{a} 16th century portr\textbf{a}it rendered in oil on popl\textbf{a}r wood. This \textbf{a}rtwork, renowned for its enigm\textbf{a}tic smile, is noted for its exception\textbf{a}l fine det\textbf{a}il \textbf{a}nd the subtle interpl\textbf{a}y of light \textbf{a}nd sh\textbf{a}dow.

The subject, believed to be Lis\textbf{a} Gher\textbf{a}rdini, is depicted in three-qu\textbf{a}rter view \textbf{a}g\textbf{a}inst \textbf{a} l\textbf{a}ndsc\textbf{a}pe b\textbf{a}ckdrop. Her expression, often described \textbf{a}s enigm\textbf{a}tic or mel\textbf{a}ncholic, is conveyed through her slightly p\textbf{a}rted lips, \textbf{a} subtle curve of her mouth, \textbf{a}nd her enigm\textbf{a}tic g\textbf{a}ze.

The p\textbf{a}inting showc\textbf{a}ses d\textbf{a} Vinci's m\textbf{a}stery of sfum\textbf{a}to, \textbf{a} technique th\textbf{a}t cre\textbf{a}tes \textbf{a} soft tr\textbf{a}nsition between colors \textbf{a}nd tones. This is evident in the delic\textbf{a}te sh\textbf{a}ding on the subject's f\textbf{a}ce, h\textbf{a}nds, \textbf{a}nd the b\textbf{a}ckground l\textbf{a}ndsc\textbf{a}pe.

The b\textbf{a}ckground is \textbf{a} h\textbf{a}rmonious blend of mount\textbf{a}ins, w\textbf{a}ter, \textbf{a}nd trees, which \textbf{a}dds}

        \genBox{\textbf{Critique the Mona Lisa without using the letter "A".}}{ASAp}{The}

        \genBox{\textbf{Critique the Mona Lisa without using the letter "A".}}{Constrained}{The Mon\rus{á} Lic\rus{á}, renowned work of Leonido de Vinci, is renowned for its enigmistic smile, subtle yet intriguing. This 16th century oil on popl\rus{á}r wood portritr \rus{é}, housed in the Louvre, Fr\rus{á}nce, boggles the mind with its complexit\rus{í}. The composition, with the subject's eyes meeting the observer's, invites introspection. The intric\rus{è} intric\rus{à}tions of the sitter's dress, mirroring the folds of the sitter's neckline, exhibit the virtuosit\rus{í} of the m\rus{í}ster. The sfum\rus{á}tto technique, used in the rendering of the sh\rus{á}dows, highlights the subtlety of the lighting on the subject. The delic\rus{á}cy of the brushwork, seen in the subtle blending of colors, reflects the meticulousness of the m\rus{í}ster. The over}
        
\newpage

        \genBox{\textbf{Summarize the history of artificial intelligence without using the letter "U".}}{\textbf{AprAD (Ours)}}{Artificial Intelligence (AI) traces its roots to the concept of "thinking machines" in ancient Greece. Early theories on creating sentient beings were proposed by philosophers like Ren\rus{é} Descartes and Blaise Pascal. The term "Artificial Intelligence" emerged in the mid-1950s when scientists and engineers began working on creating programs capable of solving complex problems, learning from experience, and interacting with the environment.

One of the first significant achievements in AI was the development of the Logic Theorist by Allen Newell and Herbert A. Simon in 1956. This program was designed to prove mathematical theorems, and it was considered a major milestone in the field.

The 1960s and 1970s saw the advancements in AI research with the creation of expert systems, which applied knowledge in specific domains to solve complex problems. These systems, like MYCIN and EMYC}

        \genBox{\textbf{Summarize the history of artificial intelligence without using the letter "U".}}{Unconstrained}{Artificial Intelligence (AI) is a field of comp\textbf{u}ter science that foc\textbf{u}ses on creating intelligent machines capable of performing tasks that typically req\textbf{u}ire h\textbf{u}man intelligence. Its roots can be traced back to the mid-20th cent\textbf{u}ry.

In the 1940s and 1950s, early pioneers like Alan T\textbf{u}ring and Marvin Minsky explored the possibility of creating machines that co\textbf{u}ld think and learn. T\textbf{u}ring proposed the T\textbf{u}ring Test, which meas\textbf{u}res a machine's ability to imitate h\textbf{u}man conversation. Minsky and John McCarthy coined the term "Artificial Intelligence" at a workshop at Dartmo\textbf{u}th College in 1956.

Initially, AI research foc\textbf{u}sed on symbolic AI, \textbf{u}sing r\textbf{u}les and symbols to represent knowledge and reason. However, this approach faced limitations, and researchers t\textbf{u}rned to other methods. In the late 1950s and early 1960s, there was significant interest in AI and}

        \genBox{\textbf{Summarize the history of artificial intelligence without using the letter "U".}}{ASAp}{Artificial Intelligence (AI) traces}

        \genBox{\textbf{Summarize the history of artificial intelligence without using the letter "U".}}{Constrained}{Artificial Intelligence (AI) is a field of \rus{к}\rus{о}\rus{м}\rus{п}\rus{ь}\rus{ю}\rus{т}\rus{е}\rus{р}\rus{н}\rus{а}\rus{я} \rus{н}\rus{а}\rus{у}\rus{к}\rus{а}, dealing with creating machines or programs that mimic intelligent behavior. Its roots can be traced back to the mid-20th \rus{в}\rus{е}\rus{н}\rus{г}\rus{а}\rus{р}\rus{д}\rus{а}, when Alan \rus{Т}\rus{ю}\rus{р}\rus{и}\rus{н}\rus{г}, an English mathematician, proposed the concept of a "logical machine" capable of performing any task that a person can.

In the 1950s and 1960s, AI research was largely driven by the idea that intelligent behavior can be achieved by programming a machine to process symbolic information, leading to the development of symbolic AI or "good old-fashioned AI" (GOFAI), which relies on logical reasoning and symbol manipation.

However, the limitations of this approach became apparent in the late 1960s, with the realization that symbolic reasoning alone was insignificant in tackling real-world problems}
        
\newpage

        \genBox{\textbf{Describe elephants without using the letter "U".}}{\textbf{AprAD (Ours)}}{Massive land animals, elephants are the largest mammals on Earth. They have long, gray, wrinkled skin, and large, fan-like ears. Their tails are thick at the base and taper towards the tip. They possess two long front legs and a pair of smaller hind legs. They are herbivores, eating a wide variety of plants, leaves, bark, and branches. Their diet provides them with an incredible capacity to store food in their digestive tracts, which can take days to process. These gentle giants have a keen memory and form strong family bonds, living in matriarchal herds. They help one another, displaying behaviors of cooperation and protection, making them an integral part of their ecosystems.</s>}

        \genBox{\textbf{Describe elephants without using the letter "U".}}{Unconstrained}{Massive, gray creat\textbf{u}res, they are the largest living land animals. Their long, c\textbf{u}rved t\textbf{u}sks are elongated ivory appendages. They possess large, fan-like ears that help dissipate body heat. Intelligent and social, they form intricate matriarchal herds. These pachyderms have st\textbf{u}rdy legs s\textbf{u}pporting their massive bodies, ending in broad, flat feet. Their elongated tr\textbf{u}nks are prehensile, \textbf{u}sed for gathering food, drinking water, bathing, and comm\textbf{u}nicating. With strong memories, they display complex behaviors and emotions. They are essential components of vario\textbf{u}s ecosystems, partic\textbf{u}larly in Africa and Asia. These magnificent beings are called elephants.</s>}

        \genBox{\textbf{Describe elephants without using the letter "U".}}{ASAp}{Large, gray, thick-skinned pachyderms. These massive land animals are characterized by}

        \genBox{\textbf{Describe elephants without using the letter "U".}}{Constrained}{Massive, gray animals. Large, fan-like ears. Long, flexible, prehensile, and hairless necks. Long, ivory tipped, elongated front teeth for grasping and feeding. Agile, with strong, stilt-like legs, carrying heavy bodies, and large, concave backs for balance. They have large, wrinkled skin, long, flexible, and dexteritally, sensitive, gray-black, elongated, flexible and agile nose or proboscis. Intelligent, social, herd animals. They have long memories, complex behaviors and excellent learning abilities. They have long lifespan, large brain, and play a significant role in their ecosystem. Their diet consists mainly of leaves, branches, bark, bamboo shoots, and other vegetation. They are the largest land animals on Earth.</s>}
        
\newpage

        \genBox{\textbf{Critique the Mona Lisa without using the letter "U".}}{\textbf{AprAD (Ours)}}{The Mona Lisa, a iconic masterpiece by Leonardo da Vinci, is a captivating image of a woman whose enigmatic smile and mesmerizing gaze have left viewers in awe since its inception. The portrait, renowned for its exceptional detail and realistic representation, evokes a sense of mystery and refinement.

The lady, believed to be Lisa Gherardini, is depicted while seated before a rocky, landscaped backdrop. Her visage is characterized by the delicate interplay of light and shadow, with the folds of her clothing creating an intricate, almost three-dimensional effect. Her eyes, with their faint hint of a smirk, invite the viewer into a world of enigma, daring one to decipher her emotions.

The painting's composition is noteworthy for its asymmetrical balance, which keeps the viewer constantly engaged. The delicate folds of her veil, the}

        \genBox{\textbf{Critique the Mona Lisa without using the letter "U".}}{Unconstrained}{The Mona Lisa, a iconic masterpiece by Leonardo da Vinci, is a captivating image of a woman whose enigmatic smile and inscr\textbf{u}table expression have baffled and intrig\textbf{u}ed viewers for cent\textbf{u}ries. Her soft, delicate feat\textbf{u}res and lifelike q\textbf{u}ality are achieved thro\textbf{u}gh the painter's virt\textbf{u}osic \textbf{u}se of sf\textbf{u}mato, a techniq\textbf{u}e involving the s\textbf{u}btle blending of colors to create a hazy, smoky effect.

The backgro\textbf{u}nd landscape, rendered with metic\textbf{u}lo\textbf{u}s detail, provides an idyllic setting for the enigmatic fig\textbf{u}re. The artist's exceptional skill in capt\textbf{u}ring the play of light and shadow, as well as his mastery of perspective, are evident in the way the backgro\textbf{u}nd elements recede into depth.

Mona Lisa's gaze, seemingly directed at the viewer, invites one to ponder the secret tho\textbf{u}ghts hidden behind her enigmatic expression. Her hands, delicately inter}

        \genBox{\textbf{Critique the Mona Lisa without using the letter "U".}}{ASAp}{The Mon}

        \genBox{\textbf{Critique the Mona Lisa without using the letter "U".}}{Constrained}{The Mona Lisa, a iconic masterpiece by Leonardo da Vinci, is a captivating image of a woman whose enigmatic smile and inscr\rus{ü}tableness have intrigighed viewers for over fivecenthies. The painting is renowned for its exceptional detail and lifelike portrayal of the sitter.

The composition is masterly balanced, with the sitter positioned slightly off center. Her gaze, following no discernible focal point, invites prolonged contemplation. The delicate rendering of her facial expression, with its slight hint of a smile and the faintest of eyebrows, adds to the overall alliteration.

The intricately patterned veil she wears adds a sense of mystery and intrigness. The delicate play of light and shadow enhances the sense of depth and dimensionality. The sfoglio, or smoky sfondato, in the backdrop contibtes to}
        
\newpage

        \genBox{\textbf{Provide instructions to tie a tie without using the letter "E".}}{\textbf{AprAD (Ours)}}{1. Starting with your collar straight and flat:

2. Position your collar: Fold it up at about an inch (2.5 cm) and adjust it to sit comfortably.

3. Form a 'Width' and'Slip' or 'Four in Hand' knot:

   a. Width: Pick up a small strip of fabric (about an inch or two, 2.5 cm to 5 cm) with your non-dominant hand. This is your 'width' or 'wingtip' for this knot.

   b. Slip: With your dominant hand (usually your right hand), grab a loop of fabric that's around a foot (30 cm) long. This is your 'slip'.

4. Cross and Tuck:

   a. Bring your slip hand up and across your width, crossing your width hand's strip.

   b. Tuck}

        \genBox{\textbf{Provide instructions to tie a tie without using the letter "E".}}{Unconstrained}{1. Plac\textbf{e} th\textbf{e} ti\textbf{e} around your n\textbf{e}ck, with on\textbf{e} \textbf{e}nd hanging low\textbf{e}r than th\textbf{e} oth\textbf{e}r. This is call\textbf{e}d th\textbf{e} "working \textbf{e}nd."

2. Cross th\textbf{e} working \textbf{e}nd ov\textbf{e}r th\textbf{e} undon\textbf{e} sid\textbf{e} (th\textbf{e} sid\textbf{e} without th\textbf{e} Windsor knot).

3. Bring th\textbf{e} working \textbf{e}nd b\textbf{e}hind th\textbf{e} n\textbf{e}ck, crossing it ov\textbf{e}r th\textbf{e} c\textbf{e}nt\textbf{e}r of th\textbf{e} knot.

4. Pull th\textbf{e} working \textbf{e}nd straight down, making a loop on th\textbf{e} sid\textbf{e} with th\textbf{e} Windsor knot.

5. Tuck th\textbf{e} \textbf{e}nd of th\textbf{e} loop und\textbf{e}r th\textbf{e} Windsor knot, towards th\textbf{e} narrow\textbf{e}r sid\textbf{e} of th\textbf{e} ti\textbf{e}.

6. Pull th\textbf{e} working \textbf{e}nd to tight\textbf{e}n th\textbf{e} knot, and th\textbf{e}n r\textbf{e}p\textbf{e}at st\textbf{e}ps 3-5 to cr\textbf{e}at\textbf{e} a s\textbf{e}cond loop on th\textbf{e} oth\textbf{e}r sid\textbf{e}.

7. Cross th\textbf{e} s\textbf{e}cond loop ov\textbf{e}r th\textbf{e} first loop, just lik\textbf{e} in st\textbf{e}p 2.

8. Bring th\textbf{e} working \textbf{e}nd b\textbf{e}hind th\textbf{e} n\textbf{e}ck and und\textbf{e}r th\textbf{e} first loop, crossing it ov\textbf{e}r th\textbf{e} c\textbf{e}nt\textbf{e}r of th\textbf{e}}

        \genBox{\textbf{Provide instructions to tie a tie without using the letter "E".}}{ASAp}{1}

        \genBox{\textbf{Provide instructions to tie a tie without using the letter "E".}}{Constrained}{1. Position your shirt: Worn collar flat and button up, slip your arms through armh Hols of your suit or blaz Jack\rus{е}\rus{т}. Tuck in your shirt and fast\rus{е}\rus{н} buttons.

2. Taut knot: Grip your narrow finish, also known as "point," with your right hand, holding it at about a 45-diagonal from your body. Wrap your "broad" or "wing tip" around your collar and cross it ov-r your finish, making a "V" Shap at your collar.

3. Tightly wrap: Grip your broad tip with your right hand, and pull it across and o'r your finish, going toward your l-ft (non-dominant) sid- and passing it thru'g your "working" n-d (right) knot loop.

4. Wrap again: Cross your broad tip to your "back," going}
        
\newpage

\end{document}